# Modelling Mixed Discrete-Continuous Domains for Planning

**Maria Fox**                                                MARIA.FOX@CIS.STRATH.AC.UK
**Derek Long**                                              DEREK.LONG@CIS.STRATH.AC.UK
*Department of Computer and Information Sciences*
*University of Strathclyde,*
*26 Richmond Street, Glasgow, G1 1XH, UK*

## Abstract

In this paper we present PDDL+, a planning domain description language for modelling mixed discrete-continuous planning domains. We describe the syntax and modelling style of PDDL+, showing that the language makes convenient the modelling of complex time-dependent effects. We provide a formal semantics for PDDL+ by mapping planning instances into constructs of hybrid automata. Using the syntax of HAs as our semantic model we construct a semantic mapping to labelled transition systems to complete the formal interpretation of PDDL+ planning instances.

An advantage of building a mapping from PDDL+ to HA theory is that it forms a bridge between the Planning and Real Time Systems research communities. One consequence is that we can expect to make use of some of the theoretical properties of HAs. For example, for a restricted class of HAs the Reachability problem (which is equivalent to Plan Existence) is decidable.

PDDL+ provides an alternative to the continuous durative action model of PDDL2.1, adding a more flexible and robust model of time-dependent behaviour.

## 1. Introduction

This paper describes PDDL+, an extension of the PDDL (McDermott & the AIPS'98 Planning Competition Committee, 1998; Fox & Long, 2003; Hoffmann & Edelkamp, 2005) family of deterministic planning modelling languages. PDDL+ is intended to support the representation of mixed discrete-continuous planning domains. PDDL was developed by McDermott (McDermott & the AIPS'98 Planning Competition Committee, 1998) as a standard modelling language for planning domains. It was later extended (Fox & Long, 2003) to allow temporal structure to be modelled under certain restricting assumptions. The resulting language, PDDL2.1, was further extended to include domain axioms and timed initial literals, resulting in PDDL2.2 (Hoffmann & Edelkamp, 2005). In PDDL2.1, *durative actions* with fixed-length duration and discrete effects can be modelled. A limited capability to model continuous change within the durative action framework is also provided.

PDDL+ provides a more flexible model of continuous change through the use of autonomous processes and events. The modelling of continuous processes has also been considered by McDermott (2005), Herrmann and Thielscher (1996), Reiter (1996), Shanahan (1990), Sandewall (1989) and others in the knowledge representation and reasoning communities, as well as by Henzinger (1996), Rasmussen, Larsen and Subramani (2004), Haroud and Faltings (1994) and others in the real time systems and constraint-reasoning communities.

The most frequently used subset of PDDL2.1 is the fragment modelling discretised change. This is the part used in the 3rd International Planning Competition and used as the basis of PDDL2.2. The continuous modelling constructs of PDDL2.1 have not been adopted by the community at large, partly because they are not considered an attractive or natural way to represent certain kinds of continuous change (McDermott, 2003a; Boddy, 2003). By wrapping up continuous change inside durative actions PDDL2.1 forces episodes of change on a variable to coincide with logical state changes. An important limitation of the continuous durative actions of PDDL2.1 is therefore that the planning agent must take full control over all change in the world, so there can be no change without direct action on the part of the agent.

The key extension that PDDL+ provides is the ability to model the interaction between the agent's behaviour and changes that are initiated by the world. Processes run over time and have a continuous effect on numeric values. They are initiated and terminated either by the direct action of the agent or by events triggered in the world. We refer to this three-part structure as the *start-process-stop* model. We make a distinction between logical and numeric state, and say that transitions between logical states are instantaneous whilst occupation of a given logical state can endure over time. This approach takes a transition system view of the modelling of change and allows a direct mapping to the languages of the real time systems community where the same modelling approach is used (Yi, Larsen, & Pettersson, 1997; Henzinger, 1996).

In this paper we provide a detailed discussion of the features of PDDL+, and the reasons for their addition. We develop a formal semantics for our primitives in terms of a formal mapping between PDDL+ and Henzinger's theory of hybrid automata (Henzinger, 1996). Henzinger provides the formal semantics of HAs by means of the *labelled transition system*. We therefore adopt the labelled transition semantics for planning instances by going through this route. We explain what it means for a plan to be valid by showing how a plan can be interpreted as an accepting run through the corresponding labelled transition system.

We note that, under certain constraints, the Plan Existence problem for PDDL+ planning instances (which corresponds to the Reachability problem for the corresponding hybrid automaton) remains decidable. We discuss these constraints and their utility in the modelling of mixed discrete-continuous planning problems.

## 2. Motivation

Many realistic contexts in which planning can be applied feature a mixture of discrete and continuous behaviours. For example, the management of a refinery (Boddy & Johnson, 2004), the start-up procedure of a chemical plant (Aylett, Soutter, Petley, Chung, & Edwards, 2001), the control of an autonomous vehicle (Léauté & Williams, 2005) and the coordination of the activities of a planetary lander (Blake et al., 2004) are problems for which reasoning about continuous change is fundamental to the planning process. These problems also contain discrete change which can be modelled through traditional planning formalisms. Such situations motivate the need to model mixed discrete-continuous domains as planning problems.

We present two motivating examples to demonstrate how discrete and continuous behaviours can interact to yield interesting planning problems. These are Boddy and Johnson's petroleum refinery domain and the battery power model of Beagle 2.

## 2.1 Petroleum refinery production planning

Boddy and Johnson (2004) describe a planning and scheduling problem arising in the management of petroleum refinement operations. The objects of this problem include materials, in the form of hydrocarbon mixtures and fractions, tanks and processing units. During the operation of the refinery the mixtures and fractions pass through a series of processing units including distillation units, desulphurisation units and cracking units. Inside these units they are converted and combined to produce desired materials and to remove waste products. Processes include the filling and emptying of tanks, which in some cases can happen simultaneously on the same tank, treatment of materials and their transfer between tanks. The continuous components of the problem include process unit control settings, flow volumes and rates, material properties and volumes and the time-dependent properties of materials being combined in tanks as a consequence of refinement operations.

An example demonstrating the utility of a continuous model arises in the construction of a gasoline blend. The success of a gasoline blend depends on the chemical balance of its constituents. Blending results from materials being pumped into and out of tanks and pipelines at rates which enable the exact quantities of the required chemical constituents to be controlled. For example, when diluting crude oil with a less sulphrous material the rate of in-flow of the diluting material, and its volume in the tank, have to be balanced by out-flow of the diluted crude oil and perhaps by other refinement operations.

Boddy and Johnson treat the problem of planning and scheduling refinery operations as an optimisation problem. Approximations based on discretisation lead to poor solutions, leading to a financial motivation for Boddy and Johnson's application. As they observe, a moderately large refinery can produce in the order of half a million barrels per day. They calculate that a 1% decrease in efficiency, resulting from approximation, could result in the loss of a quarter of a million dollars per day. The more accurate the model of the continuous dynamics the more efficient and cost-effective the refinery.

Boddy and Johnson's planning and scheduling approach is based on dynamic constraint satisfaction involving continuous, and non-linear, constraints. A domain-specific solver was constructed, demonstrating that direct handling of continuous problem components can be realistic. Boddy and Johnson describe applying their solver to a real problem involving 18,000 continuous constraints including 2,700 quadratic constraints, 14,000 continuous variables and around 40 discrete decisions (Lamba, Dietz, Johnson, & Boddy, 2003; Boddy & Johnson, 2002). It is interesting to observe that this scale of problem is solvable, to optimality, with reasonable computational effort.

## 2.2 Planning Activities for a Planetary Lander

Beagle 2, the ill-fated probe intended for the surface of Mars, was designed to operate within tight resource constraints. The constraint on payload mass, the desire to maximise science return and the rigours of the hostile Martian environment combine to make it essential to squeeze high performance from the limited energy and time available during its mission.

One of the tightest constraints on operations is that of energy. On Beagle 2, energy was stored in a battery, recharged from solar power and consumed by instruments, the on-board processor, communications equipment and a heater required to protect sensitive components from the extreme cold over Martian nights. These features of Beagle 2 are common to all deep space planetary landers.

The performance of the battery and the solar panels are both subject to variations due to ageing, atmospheric dust conditions and temperature. Nevertheless, with long periods between communication windows, a lander can only achieve dense scientific data-gathering if its activities are carefully planned and this planning must be performed against a nominal model of the behaviour of battery, solar panels and instruments. The state of charge of the battery of the lander falls within an envelope defined by the maximum level of the capacity of the battery and the minimum level dictated by the safety requirements of the lander. This safety requirement ensures there is enough power at nightfall to power the heater through night operations and to achieve the next communications session.

All operations change the state of battery charge, causing it to follow a continuous curve within this envelope. In order to achieve a dense performance, the operations of the lander must be pushed into the envelope as tightly as possible. The equations that govern the physical behaviour of the energy curve are complex, but an approximation of them is possible that is both tractable and more accurate than a discretised model of the curve would be. As in the refinery domain, any approximation has a cost: the coarser the approximation of the model, the less accurately it is possible to determine the limits of the performance of a plan.

In this paper we refer to a simplified model of this domain, which we call the Planetary Lander Domain. The details of this model are presented in Appendix C, and discussed in Section 4.3.

## 2.3 Remarks

In these two examples plans must interact with the background continuous behaviours that are triggered by the world. In the refinery domain concurrent episodes of continuous change (such as the filling and emptying of a tank) affect the same variable (such as the sulphur content of the crude oil in the tank), and the flow into and out of the tank must be carefully controlled to achieve a mixture with the right chemical composition. In the Beagle 2 domain the power generation and consumption processes act concurrently on the power supply in a way that must be controlled to avoid the supply dropping below the critical minimal threshold. In both domains the continuous processes are subject to discontinuous first derivative effects, resulting from events being triggered, actions being executed or processes interacting. When events trigger the discontinuities might not coincide with the end-points of actions. A planner needs an explicit model of how such events might be triggered in order to be able to reason about their effects.

We argue that discretisation represents an inappropriate simplification of these domains, and that adequate modelling of the continuous dynamics is necessary to capture their critical features for planning.

## 3. Layout of the Paper

In Section 4 we explain how PDDL+ builds on the foundations of the PDDL family of languages. We describe the syntactic elements that are new to PDDL+ and we remind the reader of the representation language used for expressing temporal plans in the family. We develop a detailed example of a domain, the battery power model of a planetary lander, in which continuous modelling is required to properly capture the behaviours with which a plan must interact. We complete this section with a formal proof showing that PDDL+ is strictly more expressive than PDDL2.1.

In Section 5 we explain why the theory of hybrid automata is relevant to our work, and we provide the key automaton constructs that we will use in the development of the semantics of PDDL+. In Section 6 we present the mapping from planning instances to HAs. In doing this we are using the syntactic constructs of the HA as our semantic model. In Section 7 we discuss the subset of HAs for which the Reachability problem is decidable, and why we might be interested in these models in the context of planning. We conclude the paper with a discussion of related work.

## 4. Formalism

In this section we present the syntactic foundations of PDDL+, clarifying how they extend the foregoing line of development of the PDDL family of languages. We rely on the definitions of the syntactic structures of PDDL2.1, which we call the Core Definitions. These were published in 2003 (Fox & Long, 2003) but we repeat them in Appendix A for ease of reference.

PDDL+ includes the timed initial literal construct of PDDL2.2 (which provides a syntactically convenient way of expressing the class of events that can be predicted from the initial state). Although derived predicates are a powerful modelling concept, they have not so far been included in PDDL+. Further work is required to explore the relationship between derived predicates and the start-process-stop model and we do not consider this further in this paper.

### 4.1 Syntactic Foundations

PDDL+ builds directly on the discrete fragment of PDDL2.1: that is, the fragment containing fixed-length durative actions. This is supplemented with the timed initial literals of PDDL2.2 (Hoffmann & Edelkamp, 2005). It introduces two new constructs: events and processes. These are represented by similar syntactic frames to actions. The elements of the formal syntax that are relevant are given below (these are to be read in conjunction with the BNF description of PDDL2.1 given in Fox & Long, 2003).

```
<structure-def>       ::=:events  <event-def>
<structure-def>       ::=:events  <process-def>
```

The following is an event from the Planetary Lander Domain. It models the transition from night to day that occurs when the clock variable `daytime` reaches zero.

```
(:event daybreak
    :parameters ()
    :precondition (and (not (day)) (>= (daytime) 0))
    :effect (day)
)
```

The BNF for an event is identical to that of actions, while for processes it is modified by allowing only a conjunction of process effects in the effects field. A process effect has the same structure as a continuous effect in PDDL2.1:

```
<process-effect>     ::=(<assign-op-t> <f-head> <f-exp-t>)
```

The following is a process taken from the Planetary Lander Domain. It describes how the battery state of charge, soc, is affected when power demand exceeds supply. The interpretation of process effects is explained in Section 4.2.

```
(:process discharging
    :parameters ()
    :precondition (> (demand) (supply))
    :effect (decrease soc (* #t (- (demand) (supply))))
)
```

We now provide the basic abstract syntactic structures that form the core of a PDDL+ planning domain and problem and for which our semantic mappings will be constructed. Core Definition 1 defines a simple planning instance in which actions are the only structures describing state change. Definition 1 extends Core Definition 1 to include events and processes. We avoid repeating the parts of the core definition that are unchanged in this extended version.

**Definition 1 Planning Instance** *A* planning instance *is defined to be a pair*

$$I = (Dom, Prob)$$

*where $Dom = (Fs, Rs, As, Es, Ps, arity)$ is a tuple consisting of finite sets of function symbols, relation symbols, actions, and a function arity mapping all of these symbols to their respective arities, as described in Core Definition 1. In addition it contains finite sets of* events *$Es$ and* processes *$Ps$.*

Ground events, $\mathcal{E}$, are defined by the obvious generalisation of Core Definition 6 which defines ground actions. The fact that events are required to have at least one numeric precondition makes them a special case of actions. The details of ground processes, $\mathcal{P}$, are given in Definition 2. Processes have continuous effects on primitive numeric expressions (PNEs). Core Definition 1 defines PNEs as ground instances of metric function expressions.

**Definition 2 Ground Process** *Each $p \in \mathcal{P}$ is a* ground process *having the following components:*

- **Name** *The process schema name together with its actual parameters.*

| Time | Action | Duration |
|------|--------|----------|
| 0.01: | *Action 1* | [13.000] |
| 0.01: | *Action 2* | |
| 0.71 | *Action 3* | |
| 0.9 | *Action 4* | |
| 15.02: | *Action 5* | [1.000] |
| 18.03: | *Action 6* | [1.000] |
| 19.51: | *Action 7* | |
| 21.04: | *Action 8* | [1.000] |

Figure 1: An example of a PDDL+ plan showing the time stamp and duration associated with each action, where applicable. Actions 2, 3, 4 and 7 are instantaneous, so have no associated duration.

- **Precondition** *This is a proposition, $Pre_p$, the atoms of which are either ground atoms in the planning domain or else comparisons between terms constructed from arithmetic operations applied to PNEs or real values.*

- **Numeric Postcondition** *The numeric postcondition is a conjunction of additive assignment propositions, $\mathrm{NP}_p$, the rvalues[1] of which are expressions that can be assumed to be of the form (\* #t exp) where exp is #t-free.*

**Definition 3 Plan** *A plan, for a planning instance with the ground action set $\mathcal{A}$, is a finite set of pairs in $\mathbb{Q}_{>0} \times \mathcal{A}$ (where $\mathbb{Q}_{>0}$ denotes the set of all positive rationals).*

The PDDL family of languages imposes a restrictive formalism for the representation of plans. In the temporal members of this family, PDDL2.1 (Fox & Long, 2003), PDDL2.2 (Hoffmann & Edelkamp, 2005) and PDDL+, plans are expressed as collections of time-stamped actions. Definition 3 makes this precise. Where actions are durative the plan also records the durations over which they must execute. Figure 1 shows an abstract example of a PDDL+ plan in which some of the actions are fixed-length durative actions (their durations are shown in square brackets after each action name). Plans do not report events or processes.

In these plans the time stamps are interpreted as the amount of time elapsed since the start of the plan, in whatever units have been used for modelling durations and time-dependent effects.

**Definition 4 Happening** *A happening is a time point at which one or more discrete changes occurs, including the activation or deactivation of one or more continuous processes. The term is used to denote the set of discrete changes associated with a single time point.*

---

1. Core Definition 3 defines rvalues to be the right-hand sides of assignment propositions.

## 4.2 Expressing Continuous Change

In PDDL2.1 the time-dependent effect of continuous change on a numeric variable is expressed by means of intervals of durative activity. Continuous effects are represented by update expressions that refer to the special variable `#t`. This variable is a syntactic device that marks the update as time-dependent. For example, consider the following two processes:

```
(:process heatwater
     :parameters ()
     :precondition (and (< (temperature) 100) (heating-on))
     :effect (increase (temperature) (* #t (heating-rate)))
)

(:process superheat
     :parameters ()
     :precondition (and (< (temperature) 100) (secondaryburner-on))
     :effect (increase (temperature) (* #t (additional-heating-rate)))
)
```

When these processes are both active (that is, when the water is heating and a secondary burner is applied and the water is not yet boiling) they lead to a combined effect equivalent to:

$$\frac{d\texttt{temperature}}{dt} = (\texttt{heating-rate}) + (\texttt{additional-heating-rate})$$

Actions that have continuous update expressions in their effects represent an increased level of modelling power over that provided by fixed length, discrete, durative actions.

In PDDL+ continuous update expressions are restricted to occur only in process effects. Actions and events, which are instantaneous, are restricted to the expression of discrete change. This introduces the three-part modelling of periods of continuous change: an action or event *starts* a period of continuous change on a numeric variable expressed by means of a *process*. An action or event finally *stops* the execution of that process and terminates its effect on the numeric variable. The goals of the plan might be achieved before an active process is stopped.

Notwithstanding the limitations of durative actions, observed by Boddy (2003) and McDermott (2003a), for modelling continuous change, the durative action model can be convenient for capturing activities that endure over time but whose internal structure is irrelevant to the plan. This includes actions whose fixed duration might depend on the values of their parameters. For example, the continuous activities of riding a bicycle (whose duration might depend on the start and destination of the ride), cleaning a window and eating a meal might be conveniently modelled using fixed-length durative actions. PDDL+ does not force the modeller to represent change at a lower level of abstraction than is required for the adequate capture of the domain. When such activities need to be modelled fixed duration actions might suffice.

The following durative action, again taken from the Planetary Lander Domain, illustrates how durative actions can be used alongside processes and events when it is unnecessary to expose the internal structure of the associated activity. In this case, the action models a preparation activity that represents pre-programmed behaviour. The constants

`partTime1` and `B-rate` are defined in the initial state so the duration and schedule of effects within the specified interval of the behaviour are known in advance of the application of the `prepareObs1` action.

```
(:durative-action prepareObs1
    :parameters ()
    :duration (= ?duration (partTime1))
    :condition (and (at start (available unit))
                    (over all (> (soc) (safelevel))))
    :effect (and      (at start (not (available unit)))
                      (at start (increase (demand) (B-rate)))
                      (at end (available unit))
                      (at end (decrease (demand) (B-rate)))
                      (at end (readyForObs1)))
)
```

## 4.3 Planetary Lander Example

We now present an example of a PDDL+ domain description, illustrating how continuous functions, driven by interacting processes, events and actions, can constrain the structure of plans. The example is based on a simplified model of a solar-powered lander. The actions of the system are durative actions that draw a fixed power throughout their operation. There are two observation actions, `observe1` and `observe2`, which observe the two different phenomena. The system must prepare for these, either by using a single long action, called `fullPrepare`, or by using two shorter actions, called `prepareObs1` and `prepareObs2`, each specific to one of the observation actions. The shorter actions both have higher power requirements over their execution than the single preparation action. The lander is required to execute both observation actions before a communication link is established (controlled by a timed initial literal), which sets a deadline on the activities.

These activities are all carried out against a background of fluctuating power supply. The lander is equipped with solar panels that generate electrical power. The generation process is governed by the position of the sun, so that at night there is no power generated, rising smoothly to a peak at midday and falling back to zero at dusk. The curve for power generation is shown in Figure 2. Two key events affect the power generation: at nightfall the generation process ends and the lander enters night operational mode. In this mode it draws a constant power requirement for a heater used to protect its instruments, in addition to any requirements for instruments. At dawn the night operations end and generation restarts. Both of these events are triggered by a simple clock that is driven by the twin processes of power generation and night operations and reset by the events.

The lander is equipped with a battery, allowing it to store electrical energy as charge. When the solar panels are producing more power than is required by the instruments of the lander, the excess is directed into recharging the battery (the `charging` process), while when the demand from instruments exceeds the solar power then the shortfall must be supplied from the battery (the `discharging` process). The charging process follows an inverse exponential function, since the rate of charging is proportional to the power devoted to charging and also proportional to the difference between the maximum and current levels of charge. Discharge occurs linearly at a rate determined by the current demands of all the lander activities. Since the solar generation process is itself a non-linear function of time during the day, the state of charge of the battery follows a complex curve with discontinuities in its rate of change caused by the instantaneous initiation or termination of the durative instrument actions. Figure 3 shows an example of a plan and the demand curve it generates compared with the supply over the same period.

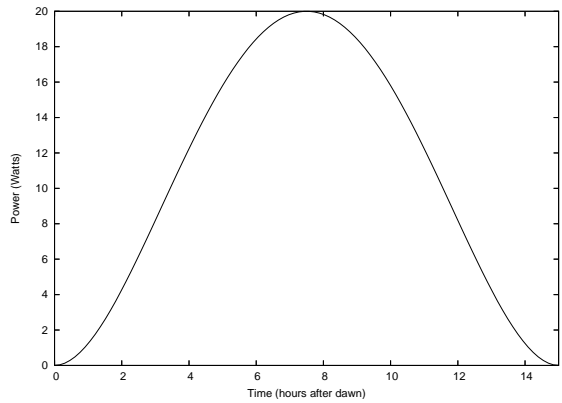

Figure 2: Graph of power generated by the solar panels.

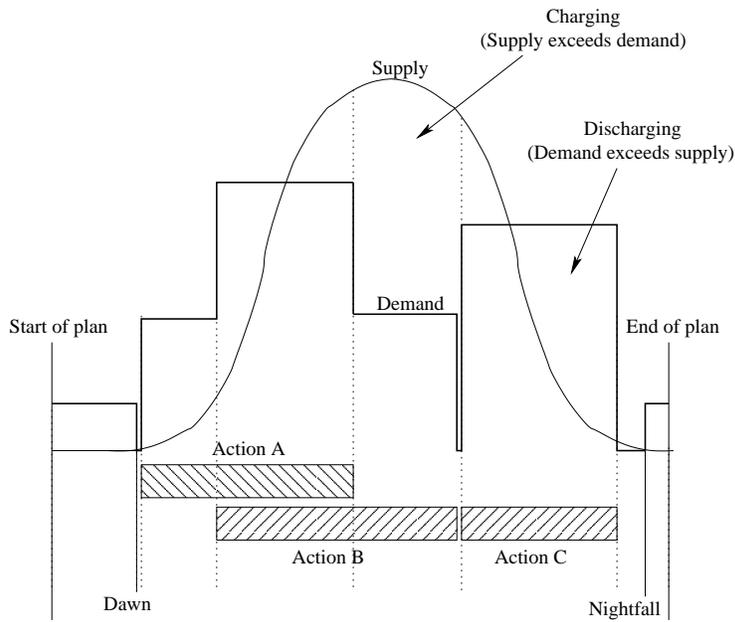

Figure 3: An abstracted example lander plan showing demand curve and supply curve over the period of execution.

Figures 5 and 6 show graphs of the battery state of charge for the two alternative plans shown in Figure 4. The plans both start an hour before dawn and the deadline is set to 10 hours later. The parameters have been set to ensure that there are 15 hours of daylight, so

```
0.1: (fullPrepare) [5]          2.6: (prepareObs1) [2]
5.2: (observe1) [2]             4.7: (observe1) [2]
7.3: (observe2) [2]             6.8: (prepareObs2) [1]
                                7.9: (observe2) [2]
```

Figure 4: Two alternative plans to complete the observations before the deadline.

the plan must complete within two hours after midday. The battery begins at 45% of fully charged.

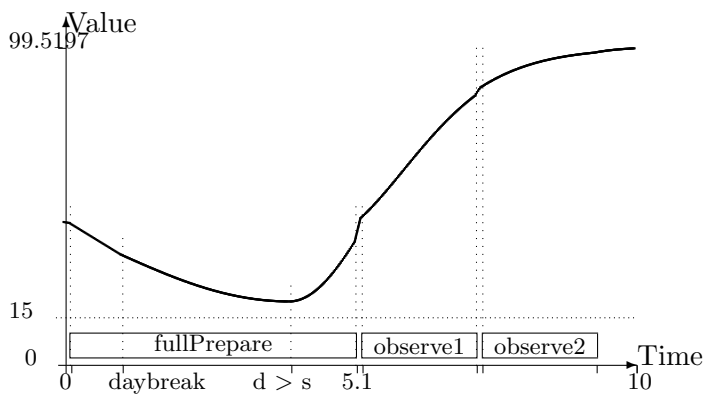

Figure 5: Graph of battery state of charge (as a percentage of full charge) for first plan. The timepoint marked $d > s$ is the first point at which demand exceeds supply, so that the battery begins to recharge. The vertical lines mark the points at which processes are affected. Where the state of charge is falling over an interval the discharge process is active and where it is rising the charge process is active.

The lander is subject to a critical constraint throughout its activities: the battery state of charge may never fall below a safety threshold. This is a typical requirement on remote systems to protect them from system failures and unexpected problems and it is intended to ensure that they will always have enough power to survive until human operators have had the opportunity to intervene. This threshold is marked in Figure 5, where it can be seen that the state of charge drops to approximately 20%. The lowest point in the graph is at a time 2.95 hours after dawn, when the solar power generation just matches the instrument demand. At this point the discharging process ends and the generation process starts. This time point does not correspond to the start or end of any of the activities of the lander and is not a point explicitly selected by the planner. It is, instead, a point defined by the intersection of two continuous functions. In order to confirm satisfaction of the constraint, that the state of charge may never fall below its safety threshold, the state of charge must be monitored throughout the activity. It is not sufficient to consider its value at only its end points, where the state of charge is well above the minimum required, since the curve might dip well below these values in the middle.

We will use this example to illustrate further points later in this paper. The complete domain description and the initial state for this problem instance can be found in Appendix C, while the two reports generated by VAL (Howey, Long, & Fox, 2004) are available in the online appendices associated with this paper.

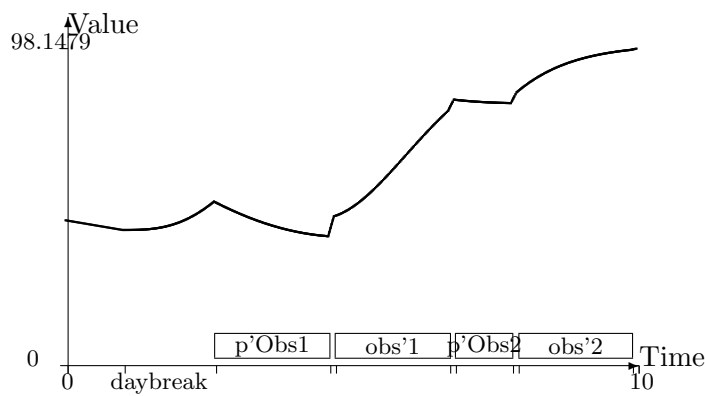

Figure 6: Graph of battery state of charge (as a percentage of full charge) for second plan. As in the previous case, discontinuities in the gradient of the state of charge correspond to points at which the charge or discharge process is changed by an action (start or end point) or event.

## 4.4 Expressive Power of PDDL+

We now consider whether PDDL+ represents a real extension to the expressive power of PDDL2.1. Of course, the fragment of PDDL2.1 that was used in the competition and has been widely used since (the fragment restricted to discrete durative actions) does not include the parts that express continuous change, and without those elements PDDL2.1 is certainly less expressive than PDDL+. In this section we discuss the differences between modelling continuous change using the continuous durative action constructs of PDDL2.1, and modelling it using the start-process-stop model.

PDDL2.1, complete with continuous durative actions, comprises a powerful modelling language. Allowing continuous effects within flexible duration actions offers an expressive combination that appears close to the processes and events of PDDL+. The essential difference between the languages arises from the separation, in PDDL+, between the changes to the world that are directly enacted by the executive and those indirect changes that are due to physical processes and their consequences.

To model physical processes and their consequences in PDDL2.1 requires the addition to the domain model of artificial actions to simulate the way in which processes and events interact with eachother and with the direct actions of the executive. For example, to force intervals to abut, so that the triggering of an event is correctly modelled, requires artificial actions that force the corresponding end points of the intervals to synchronise. These actions must be applied by the planner, since there is no other dynamic to force indirect events to coincide in the way that they would coincide in nature. In earlier work (Fox & Long, 2004) we show how *clips* can be constructed in PDDL2.1 and used to achieve this effect. Clips prevent time passing between the end points of actions modelling the background behaviour of the world. The example shown in Figure 7 illustrates how clips can be used to model interacting continuous effects in a PDDL2.1 representation of the Planetary Lander Domain.

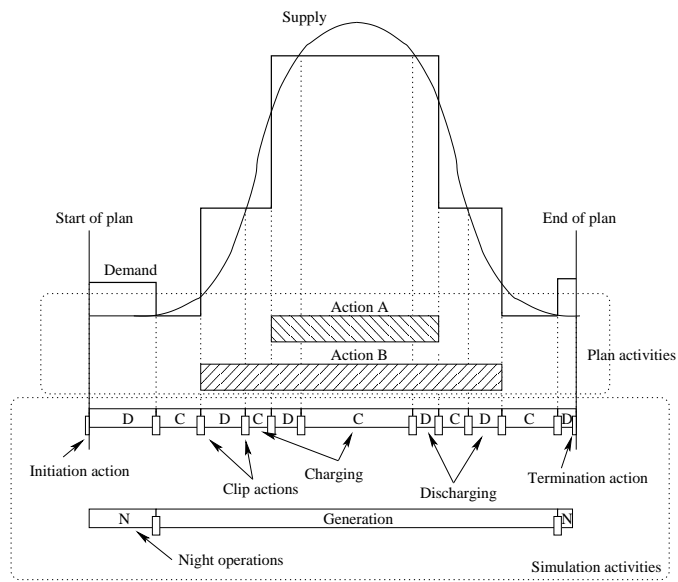

Figure 7: An illustration of the structure of the simulation activities required to model a simple PDDL+ plan in PDDL2.1. A and B are actions, C and D are the `charging` and `discharging` processes, respectively.

The example shows two activities executing against a backdrop of continuous charging and discharging of a battery. The bell-shaped curve represents the solar power production, which starts at daybreak, reaches its peak at midday and then drops off to zero at nightfall. Two concurrent power-consuming activities, A and B, are executing during the daylight hours. The stepped curve shows their (cumulative) power requirements. A PDDL+ representation of this plan would contain just the two actions A and B — the processes and events governing power consumption and production would be triggered autonomously and would not be explicit in the plan. By contrast, a PDDL2.1 representation of the same plan would contain durative actions for each of the episodes of charge, C, and discharge, D, which need to be precisely positioned (using clips) with respect to the two activities A and B. Clips are required because actions C and D do not have fixed durations so they have to be joined together to force them to respect the underlying timeline. The two dotted rectangles of the figure depict a PDDL2.1 plan containing 28 action instances in addition to A and B. Of these, there are four points simulating the intersections between the supply and the demand curves that do not correspond to the end points of actions A and B. A planner using a PDDL2.1 model is forced to construct these points in its simulation of the background behaviours. All of the necessary clip actions would be explicit in the plan.

As can be seen, the construction of an accurate simulation in PDDL2.1 is far from trivial. Indeed, although this is not an issue that we highlight in the example, there are cases where no simulation can be constructed to be consistent with the use of $\epsilon$-separation for interfering action effects. This occurs when events or process interactions occur with arbitrarily small temporal separations. Even where simulations can be constructed, the lack of distinction between direct and indirect causes of change means that a planner is forced to construct the simulated process and event sequences as though they are part of the plan it is constructing. This means that the planner is required to consider the simulation components as though they are choice points in the plan construction, leading to a combinatorial blow up in the cost of constructing plans.

The explicit distinction between actions and events yields a compact plan representation in PDDL+. A PDDL2.1 plan, using a simulation of the events and processes, would contain explicit representations of every happening in the execution trace of the PDDL+ plan. The fact that the PDDL2.1 plan represents a form of constructive proof of the existence of an execution trace of the PDDL+ plan is one way to understand Theorem 1 below: the work in validating a PDDL+ plan is that required to construct the proof that a PDDL2.1 plan would have to supply explicitly.

By distinguishing between the direct action of the executive and the continuous behaviours of the physical world we facilitate a decomposition of the planning problem into its discrete and continuous components. This decomposition admits the use of hybrid reasoning techniques, including Mixed Integer Non-Linear Programming (MINLP) (Grossmann, 2002), Benders Decomposition (Benders, 1962), Branch-and-Bound approaches that relax the discrete components of the domain into continuous representations (Androulakis, 2001), and other such techniques that have proved promising in mixed discrete-continuous problem-solving (Wu & Chow, 1995). By contrast, trying to treat a hybrid problem using purely discrete reasoning techniques seems likely to result in an unmanageable combinatorial explosion. Of course, the trade-offs cannot be fully understood until planners exist for tackling mixed discrete-continuous domains featuring complex non-linear change.

We now prove that PDDL+ has a formally greater expressive power than PDDL2.1.

**Theorem 1** PDDL+ *is strictly more expressive than* PDDL2.1.

**Proof:** We demonstrate this by showing that we can encode the computation of an arbitrary register machine (RM) in the language of PDDL+. The instructions of the RM are encoded as PDDL+ events and the correct execution of a plan can be made to depend on the termination of the corresponding RM program. This means that the general plan validation problem for PDDL+ plans is undecidable, while for PDDL2.1 plans it is decidable. This is because PDDL2.1 plans explicitly list all the points in a plan at which a state transition occurs (as actions) and these can be checked for validity by simulated execution. In contrast, a PDDL+ plan leaves events implicit, so a plan cannot be tested without identifying the events that are triggered and confirming their outcomes.

To simulate an arbitrary RM program, we need an action that will initiate execution of the program:

```
(:action start
 :parameters ()
 :precondition ()
 :effect (started))
```

Now we construct a family of events that simulate execution of the program. We use an encoding of a register machine with three instructions: `inc(j,k)` which increments register $j$ and then jumps to instruction $k$, `dec(j,k,z)`, which tests register $j$ and jumps to instruction $z$ if it is zero and otherwise decrements it and jumps to instruction $k$, and `HALT` which terminates the program. We assume that the instructions are labelled $0, \ldots, n$ and the registers used are labelled $0, \ldots, m$. We also assume that instruction 0 is the start of the program.

```
(:event beginExection
 :parameters ()
 :precondition (started)
 :effect      (and (not (started))
                   (in_0)))
```

For an instruction of the form: $l$: `inc(j,k)` where $l$ is the label, we construct:

```
(:event do_l
 :parameters ()
 :precondition (in_l)
 :effect      (and (not (in_l))
                   (in_k)
                   (increase (reg_j) 1)))
```

For an instruction of the form: $l$: `dec(j,k,z)` we construct:

```
(:event do_l
 :parameters ()
 :precondition (in_l)
 :effect      (and (not (in_l))
                   (when (= (reg_j) 0) (in_z))
                   (when (> (reg_j) 0) (and (decrease (reg_j) 1)
                                            (in_k)))))
```

Finally, for an instruction $l$: `HALT` we have:

```
(:event do_l
 :parameters ()
 :precondition (in_l)
 :effect      (and (not (in_l))
                   (halted)))
```

We now create an initial state in which registers `reg_0...reg_m` are all initialised to 0 and the goal `halted`. It is now apparent that the plan:

```
    1:  (beginExecution)
```

is valid if and only if the computation of the embedded RM halts. Therefore, a general plan validation system for PDDL+ would have to be able to solve the halting problem.

$\square$

Theorem 1 is a formal demonstration of the increase in expressive power offered by PDDL+. It depends on the fact that a PDDL+ plan is defined to exclude explicit indication of events and processes that are triggered during the execution of the plan. It might be argued that this is an artificial problem, but there are two points to consider. Firstly, by avoiding the requirement that events and processes be captured explicitly in the plan we remain agnostic about the nature of the reasoning that a planner might perform about these phenomena. It might be that a planner can synthesise an approximation of a continuous process that simplifies the reasoning and is sufficiently accurate to allow it to place its actions around the process behaviour, but that would be insufficient to determine the precise moments at which the process triggers events. Secondly, a planner might be able to determine that some collection of processes and events is irrelevant to the valid execution of a plan it has constructed to solve a problem, even though it is apparent that some pattern of processes and events will be triggered during the execution of the plan. In this case, the requirement that the plan correctly and explicitly captures all of this background activity is an unreasonable additional demand.

The undecidability of the PDDL+ validation problem need not be confronted in practice. If certain restrictions are imposed (no cascading events, functions restricted to polynomials and some exponentials), which do not undermine the ability to capture realistic domains, the processes and events underlying a PDDL+ plan can be efficiently simulated using well-known numerical methods. In (Fox, Howey, & Long, 2006) we show how numerical simulation is achieved in the PDDL+ plan validator, VAL. A validation procedure must simulate these processes and events to ensure that critical values remain in acceptable ranges throughout the plan (and satisfy the conditions of planned actions). The restrictions that it is sensible to apply, in particular to sequences of events that may be triggered at the same time point, but also to the forms of continuous functions that arise in a domain, do not prevent us from achieving close approximations of realistic behaviours. Boddy and Johnson (2004) and Hofmann and Williams (2006) use linear and quadratic approximations to model complex non-linear functions. We use a quartic approximation and an inverse exponential function to represent the power dynamics in our own model of the planetary lander (see Appendix C).

In practice, although there is a formal separation between the expressive power of PDDL+ and PDDL2.1, the conceptual separation between the activities of the executive and those of the world is the most important feature of PDDL+.

We now present a simple family of domains to illustrate that PDDL2.2 encodings grow larger than PDDL+ domains encoding equivalent behaviours. The difference arises from the fact that durative actions encapsulate not only the way in which a process starts, but also the way it concludes. This means that in domains where there is a significant choice of different ways to start and end a process, the PDDL2.1 encoding expands faster than the corresponding PDDL+ encoding. Consider the PDDL+ domain containing the following action and process schemas:

```
(:action A_i
    :parameters ()
    :precondition (and (not (started)) (a_i))
    :effect (and (started) (not (a_i)) (assign (dur) dA_i))
)

(:action B_j
    :parameters ()
    :precondition (and (b_j) (= (C) (* (dur) dB_j))
```

```
    :effect (and (not (started)) (done))
)

(:process P
    :parameters ()
    :precondition (started)
    :effect (increase (C) (* #t 1))
)
```

The action schemas are families $A_i$ and $B_j$, indexed by $i$ and $j$ which take values $i \in \{1, ..., n\}$ and $j \in \{1, ..., m\}$ respectively. The values $dA_i$ and $dB_j$ are (different) action-instance-dependent constants. Any plan starting in an initial state $\{(a_x), (b_y)\}$, with $C = 0$, that achieves *done*, must contain the actions $A_x, B_y$, separated by exactly $dA_x.dB_y$. To encode an equivalent durative action model requires an action schema:

```
(:durative-action AB_{i,j}
    :parameters ()
    :duration (= ?duration (* dA_i  dB_j))
    :condition (and (at start (a_i)) (at end (b_j)))
    :effect (and (at start (not (a_i))) (at end (done)))
)
```

As can be seen, the size of the encoding of the family of PDDL+ domains grows as $O(n + m)$, while the corresponding size of the PDDL2.2 encodings grows as $O(n.m)$. The need to couple each possible initiation of the process with each possible conclusion to the process leads to this multiplicative growth. Reification of the propositions in the durative action encoding can be used to reduce the encoding to an $O(n + m)$ encoding, but the ground action set continues to grow as $O(n.m)$ compared with the $O(n + m)$ growth of the ground actions and processes for the PDDL+ model.

It is clear that it is easier to build the plan given the $O(n.m)$ encoding, because this provides ready-made solutions to the problem. However, the trade-off to be explored lies in how large a representation can be tolerated to obtain this advantage in general. It is always possible to compile parts of the solution to a problem into the problem representation, but the price that is paid is in the size of the encoding and the effort required to construct it. On this basis we argue that a compact representation is preferable. The example we have presented is an artificial example demonstrating a theoretical difference in the expressive powers of PDDL2.2 and PDDL+. It remains to be seen whether this phenomenon arises in practice in realistic domains.

## 5. PDDL+ and Hybrid Automata

In this section we discuss the role of Hybrid Automata in relation to PDDL+. We motivate our interest in Hybrid Automata and then proceed to describe them in more detail.

### 5.1 The Relevance of HA Theory

Researchers concerned with the modelling of real-time systems have developed techniques for modelling and reasoning about mixed discrete-continuous systems (Yi et al., 1997; Henzinger, Ho, & Wong-Toi, 1995; Rasmussen et al., 2004). These techniques have become well-established. The theory of hybrid automata (Henzinger, 1996; Gupta, Henziner, & Jagadeesan, 1997; Henzinger & Raskin, 2000), which has been a focus of interest in the model-checking community for some years, provides an underlying theoretical basis for such work. As discussed in Section 2, the central motivation for the extensions introduced in PDDL+ is to enable the representation of mixed discrete-continuous domains. Therefore, the theory of hybrid automata provides an ideal formal basis for the development of a semantics for PDDL+.

Henzinger (1996) describes a digital controller of an analogue plant as a paradigmatic example of a mixed discrete-continuous system. The discrete states (control modes) and dynamics (control switches) of the controller are modelled by the vertices and edges of a graph. The continuous states and dynamics of the plant are modelled by vectors of real numbers and differential equations. The behaviour of the plant depends on the state of the controller, and vice versa: when the controller switches between modes it can update the variables that describe the continuous behaviour of the plant and hence bring about discrete changes to the state of the plant. A continuous change in the state of the plant can affect invariant conditions on the control mode of the controller and result in a control switch.

In a similar way, PDDL+ distinguishes processes, responsible for continuous change, from events and actions, responsible for discrete change. Further, the constraint in PDDL+, that numeric values only appear as the values of functions whose arguments are drawn from finite domains, corresponds to the requirement made in hybrid automata that the dimension of the automaton be finite.

An important contribution of our work is to demonstrate that PDDL+ can support succinct encodings of deterministic hybrid automata for use in planning. We expect that both the formal (semantics and formal properties) and practical (model-checking techniques) results in Hybrid Automata theory will be able to be exploited by the planning community in addressing the problem of planning for discrete-continuous planning domains. Indeed, some cross-fertilisation is already beginning (Dierks, 2005; Rasmussen et al., 2004; Edelkamp, 2003).

### 5.2 Hybrid Automata

We now present the relevant definition of a Hybrid Automaton from Henzinger's theory (Henzinger, 1996) that will be used in the construction of our formal semantics for PDDL+ planning domains.

**Definition 5 Hybrid Automaton** *A Hybrid Automaton $H$ consists of the following components:*

- **Variables.** *A finite set $X = \{x_1, \ldots, x_n\}$ of real-valued variables. The number $n$ is called the* dimension *of $H$. We write $\dot{X}$ for the set $\{\dot{x}_1, \ldots, \dot{x}_n\}$ of dotted variables,*

> representing first derivatives during continuous change, and $X'$ for the set $\{x'_1, \ldots, x'_n\}$ of primed variables, representing values at the conclusion of discrete change.

- **Control Graph.** *A finite directed graph* $\langle V, E \rangle$. *The vertices in $V$ are* control modes. *The edges in $E$ are* control switches.

- **Initial, invariant and flow conditions.** *Three vertex labelling functions, init, inv, and flow, that assign to each control mode $v \in V$ three predicates. Each initial condition init(v) is a predicate whose free variables are from $X$. Each invariant condition inv(v) is a predicate whose free variables are from $X$. Each flow condition flow(v) is a predicate whose free variables are from $X \cup \dot{X}$.*

- **Jump conditions.** *An edge labelling function* jump *that assigns to each control switch $e \in E$ a predicate. Each jump condition jump(e) is a predicate whose free variables are from $X \cup X'$.*

- **H-Events.** *A finite set $\Sigma$ of h-events and a function, hevent* : $E \rightarrow \Sigma$, *that assigns to each control switch an h-event.*

An *h-event* is referred to by Henzinger as an *event*, but we have changed the name to avoid terminological confusion with events in PDDL+.

Figure 8 shows a very simple dynamic system expressed as a hybrid automaton.

The initial, jump and flow conditions are not necessarily satisfied by unique valuations. When multiple valuations satisfy these conditions it is possible for the behaviour of the automaton to be non-deterministic.

It should be observed that Henzinger's model needs to be extended, in a simple way, to include the undefined value, $\perp$, for real-valued variables. This is because PDDL+ states can contain unassigned real-valued variables, as shown in Core Definition 2 which defines the metric valuation of a state. The role of this value is to allow for the situations in which a metric fluent is created in a domain, but is not given an initial value. Such a fluent will be given the undefined value and attempts to inspect it before it is assigned a value will be considered to yield an error. This introduces no semantic difficulties and we have left the details of this modification implicit.

Just as the input language of a finite automaton can be defined to be the set of all sequences of symbols from its alphabet, it is possible to define the input language of a Hybrid Automaton. In this case, the elements of the language are called traces:

**Definition 6 Trace** *Given a Hybrid Automaton, $H$, with h-event set $\Sigma$, a trace for $H$ is an element of the language* $(\mathbb{R}_{\geq 0} \cup \Sigma)^*$.

Informally, a trace consists of a sequence of h-events interleaved with real values corresponding to time periods during which no h-events are applied. The time at which each h-event is applied is readily determined by summing the values of the time periods in the sequence up to the point at which the h-event appears: h-events take no time to execute. Note that this definition does not require that the trace is accepted by the Hybrid Automaton: this is a property of traces we consider in Section 6.4.

A further minor point to note is that the definition of a trace allows traces where the first transition occurs at time 0. Our convention in the semantics of PDDL2.1 is to forbid actions to occur at time 0. The reason for this is discussed in (Fox & Long, 2003), but, briefly, in order to be consistent with the model in which states hold over an interval that is closed on the left and open on the right, the initial state (which holds at time 0) must persist for a non-zero interval. As a consequence, we will not be interested in traces that have action transitions at time 0.

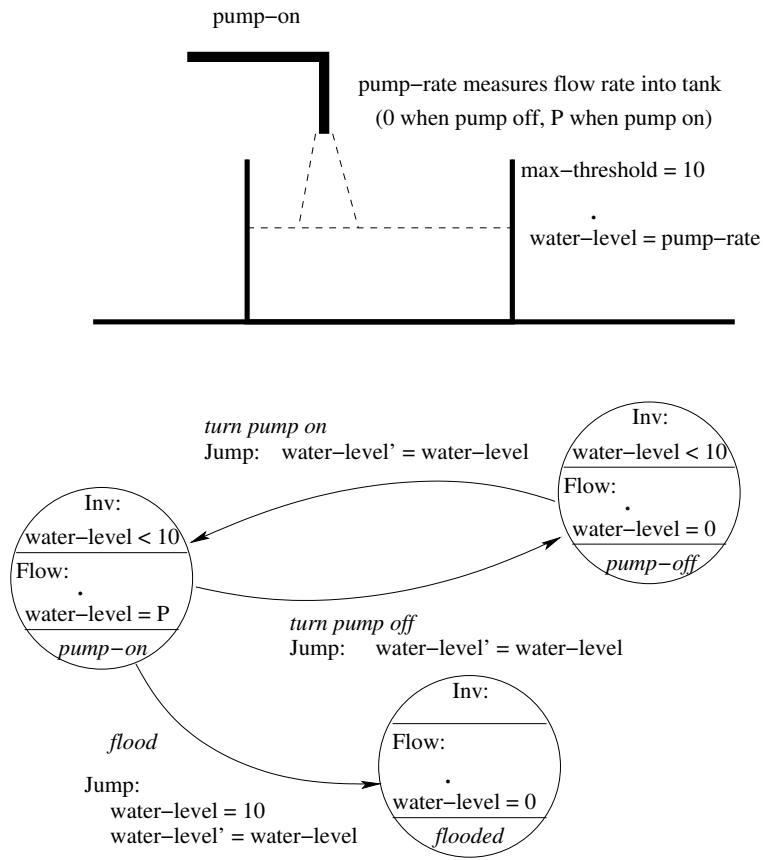

Figure 8: A simple tank-filling situation modelled as a hybrid automaton. This has three control modes and three control switches. The control switch *flood* has a jump condition which requires the level to exceed the bath capacity. Flow conditions govern the change in water level.

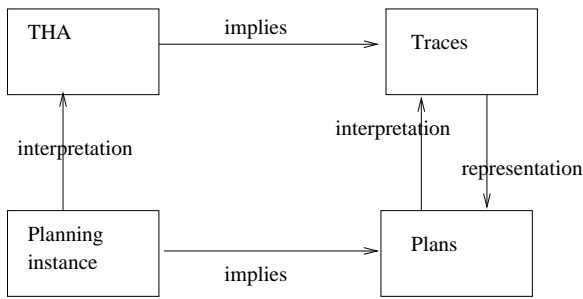

Figure 9: The semantic mapping between plans and traces

## 6. Semantics

In this section we present the semantics of PDDL+. We begin by explaining our approach and then proceed to develop the semantics incrementally.

### 6.1 Semantics of PDDL+

We present the semantics in two stages. In Section 6.2 we give the semantics of planning instances in terms of Hybrid Automata, by defining a formal mapping from planning constructs to constructs of the corresponding automata. Figure 9 illustrates the syntactic relationship between a PDDL+ planning instance and the plans that it implies, and the semantic relationships both between the PDDL+ instance and the corresponding Hybrid Automaton and between the plans implied by the model and the traces implied by the automaton. The PDDL+ instance and its plans are syntactic constructs for which the HA and its traces provide a formal semantics. We show that, whilst plans can be interpreted as traces, traces can be represented as plans by means of the abstraction of events appearing in the traces. The figure represents the first stage in the development of our formalisation. We then summarise, in Section 6.3, Henzinger's interpretation of Hybrid Automata in terms of labelled transition systems and the accompanying transition semantics. We use this semantic step as the basis of the second stage of our formalism, as is shown in Figure 13.

The important distinction that we make in our planning models between actions and events requires us to introduce a *time-slip* monitoring process (explained below), which is used to ensure events are executed immediately when their preconditions are satisfied.

In Core Definition 4 we define how PNEs are mapped to a vector of position-indexed variables, $\vec{X} = \langle X_1, ..., X_n \rangle$. The purpose of this mapping is to allow us to define and manipulate the entire collection of PNEs in a consistent way. The collection is given a valuation in a state as shown in Core Definition 2 where the logical and metric components of a state are identified. The updating function defined in Core Definition 8 specifies the relationship that must hold between the valuations of the PNEs before and after the application of an action. Normalisation of expressions that use PNEs involves replacing each of the PNEs with its corresponding position-indexed variable denoting its position in the valuation held within a state. This is performed by the semantic function $\mathcal{N}$. Update expressions are constructed using the "primed" form of the position-indexed variable for the lvalue[2] of an effect, to distinguish the pre- and post-condition values of each variable. In mapping planning instances to Hybrid Automata we make use of this collection of position-indexed variables to form the set of metric variables of the constructed automaton. In Definition 5, Henzinger uses the names $X$, $X'$ and $\dot{X}$ for the vectors of variables, the post-condition variables following discrete updates and the derivatives of the variables during continuous change, respectively. In order to reduce the potential for confusion in the following, we note that we use $X$, $X'$ and $\dot{X}$ as Henzinger does, $X_1, \ldots, X_n$ as the names of the position-indexed variables for the planning instance being interpreted and $X_{n+1}$ as an extra variable used to represent time-slip.

## 6.2 The Semantics of a Planning Instance

In the following we present the semantics of a planning instance in stages in order to facilitate understanding. We begin with the definition of a *uniprocess planning instance* and an *event-free uniprocess planning instance*. We then introduce the general concept of a planning instance. These definitions rely on the concept of *relevance* of actions, events and processes, which we now present.

The following definition uses the interpretation of preconditions defined in Core Definition 9. This core definition explains how, given a proposition $P$ and a logical state $s$, the truth of the proposition is determined. $Num(s, P)$ is a predicate over the PNEs in the domain. As explained in Core Definition 9, to determine the truth of a proposition in a state, with respect to a vector of numeric values $\vec{x}$, the formal numeric parameters of the proposition are substituted with the values in $\vec{x}$ and the resulting proposition is evaluated in the logical state. The purpose of Definition 7 is to identify actions, events or processes that *could* be applicable in a given logical state, if the values of the metric fluents are appropriate to satisfy their preconditions.

**Definition 7 Relevance of Actions, Events and Processes** *A ground action a (event e, or process p) of a planning instance I of dimension n, is* relevant *in the logical state s if there is some value $\vec{x} \in \mathbb{R}^n$ such that $Num(s, Pre_a)(\vec{x})$ ($Num(s, Pre_e)(\vec{x})$ or $Num(s, Pre_p)(\vec{x})$, respectively).*

$\mathcal{P}_s$ ($\mathcal{E}_s$) *is the set of all ground processes (events) that are* relevant *in state s.*

In general, an action, event or process that is relevant in a particular (logical) state might not actually become applicable, since the valuations of the numeric state that arise while the system is in the logical state might not include any of those that satisfy the preconditions of the corresponding transition.

In the construction of a HA, in the mappings described below, the vertices of its control graph are subsets of ground atoms and are therefore equivalent to logical states. We use the variable $v$ to denote a vertex and $\mathcal{P}_v$ ($\mathcal{E}_v$) to denote the processes (events) relevant in the corresponding logical state.

---

2. The *lvalue* in an update expression is the variable on the left of the expression, to which the value of the expression on the right is assigned.

In any given logical state a subset of the relevant processes will be active, according to the precise valuation of the metric fluents in the current state. We first consider the restricted case in which no more than one active process affects the value of each variable at any time, which we call a *uniprocess planning instance*. The proposition *unary-context-flow($i, \pi$)*, defined below, is used to describe the effects on the $i$th variable of the process $\pi$, when it is active. We later extend this definition to the concurrent case in which multiple processes may contribute to the behaviour of a variable.

**Definition 8 Uniprocess Planning Instance** *A planning instance is a* uniprocess planning instance *if, for each metric variable, $X_i$, the set of processes that can affect the value of $X_i$ relevant in state $v$, denoted $\mathcal{P}_v|_{X_i}$, contains processes whose preconditions are pairwise mutually exclusive. That is, for $\pi_1, \pi_2 \in \mathcal{P}_v|_{X_i}$ ($\pi_1 \neq \pi_2$) there is no numeric state such that $\mathcal{N}(Pre_{\pi_1}) \wedge \mathcal{N}(Pre_{\pi_2})$.*

**Definition 9 Unary-context-flow** *If $v$ is a logical state for a uniprocess planning instance $I$ of dimension $n$ and $\pi \in \mathcal{P}_v|_{X_i}$ then the* unary-context-flow *proposition is defined as follows. Let the effect of $\pi$ on $X_i$ take the form* (increase $X_i$ (* #t $Q_i$)) *for some expression $Q_i$.*

$$unary\text{-}context\text{-}flow_v(i, \pi) = (\mathcal{N}(Pre_\pi) \rightarrow \dot{X}_i = Q_i)$$

We begin by presenting the semantics of a *event-free uniprocess planning instance* — that is, a uniprocess planning instance that contains no events. We then extend our definition to include events, and present the semantics of a *uniprocess planning instance*. Finally we introduce concurrent process effects and the definition of the semantics of a *planning instance*.

**Definition 10 Semantics of an Event-free Uniprocess Planning Instance** *An event-free uniprocess planning instance, $I = (Dom, Prob)$, is interpreted as a Hybrid Automaton, $H_I$, as follows:*

- **Variables.** *The variables of $H_I$ are $X = \{X_1, \ldots, X_n\}$, where $n$ is the dimension of the planning problem.*

- **Control Graph.** *The set of vertices $V$ is formed from all subsets of the ground atoms of the planning instance. The set of edges in $E$ contains an edge $e$ between $v$ and $v'$ iff there is an action, $a$, relevant in $v$ and:*

$$v' = (v - Del_a) \cup Add_a$$

*The action $a$ is associated with the edge $e$.*

- **Initial, invariant and flow conditions.** *The vertex labelling function* init *is defined as:*

$$init(v) = \begin{cases} false & if\ v \neq Init_{logical} \\ \mathcal{N}(Init_{numeric}) \wedge \bigwedge \{X_i = \perp\ |X_i \notin \mathcal{N}(Init_{numeric})\} & otherwise \end{cases}$$

*The vertex labelling function* inv *is the proposition True.*

*The vertex labelling function* flow *is defined:*

$$flow(v) = (\bigwedge_{i=1}^{n} \{ \bigwedge_{\pi \in \mathcal{P}_v|_{X_i}} \text{unary-context-flow}_v(i, \pi) \wedge (\bigwedge_{\pi \in \mathcal{P}_v|_{X_i}} \neg\mathcal{N}(Pre_\pi) \rightarrow \dot{X}_i = 0) \})$$

- **Jump conditions.** *The edge labelling function* jump *is defined as follows. Given an edge e from vertex v, associated with an action a:*

$$jump(e) = \mathcal{N}(Pre_a) \wedge UF_a(X) = X'$$

  *where $UF_a$ is the updating function for action a, as defined in Core Definition 8 which specifies the relationship that must hold between the valuations of the PNEs before and after the application of an action.*

- **H-events.** $\Sigma$ *is the set of all names of all ground actions. The edge labelling function* hevent : $E \rightarrow \Sigma$ *assigns to each edge the name of the action associated with the edge.*

The flow condition states that, for each variable, if the precondition of one of the processes that could affect it is true then that process defines its rate of change (through the unary-context-flow proposition), or else, if none of the process preconditions is satisfied then the rate of change of that variable is zero.

To illustrate this construction we now present a simple example. The planetary lander domain requires events and so cannot be used as an example of the event-free model. Consider the PDDL+ domain containing just the following actions and process:

```
(:action startEngine
 :precondition (stopped)
 :effect (and (not (stopped))
              (running))
)

(:action accelerate
 :precondition (running)
 :effect (increase (a) 1)
)

(:action decelerate
 :precondition (running)
 :effect (decrease (a) 1)
)

(:action stop
 :precondition (and (= (v) 0) (running))
 :effect (and (not (running))
              (stopped)
              (assign (a) 0))
)
```

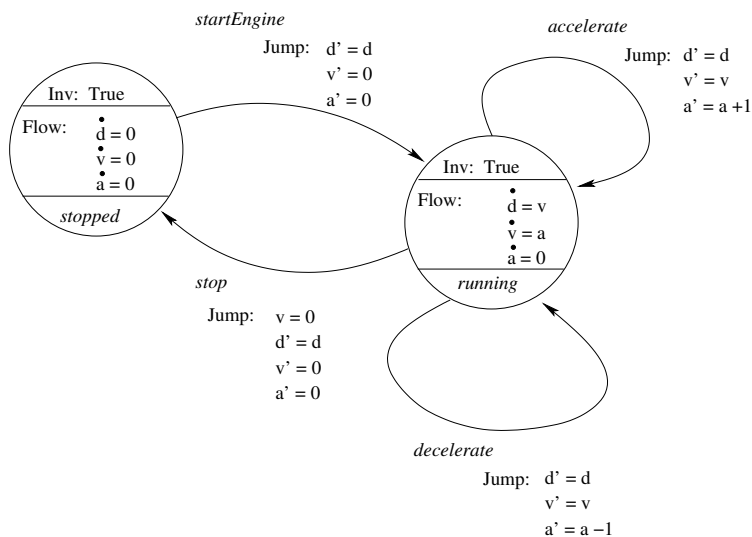

Figure 10: A hybrid automaton constructed by the translation of an event-free uniprocess planning instance. We ignore the *init* function, which simply asserts the appropriate initial state for a particular problem instance.

```
(:process moving
 :precondition (running)
 :effect (and (increase (d) (* #t (v)))
              (increase (v) (* #t (a))))
)
```

The translation process described in Definition 10 leads to the hybrid automaton shown in Figure 10. As can be seen, there are two vertices, corresponding to the logical states {*stopped*} and {*running*}. The four actions translate into edges, each edge linking a vertex in which an action is relevant to one in which its logical effects have been enacted. The metric effects of actions are encoded in the jump conditions associated with an edge, using the convention that the primed versions of the variables refer to the state following the transition. Note that variables that are not explicitly affected by an action are constrained to take the same value after the transition as they did before the transition: this is the metric equivalent of the STRIPS assumption. The *stop* action has a metric precondition and this is expressed in the jump condition of the transition, requiring that the velocity variable, $v$, be zero for this transition. In the stopped state no process can affect the variables, so the flow conditions simply assert that the variables have a zero rate of change. In the *running* state the *moving* process is relevant — indeed, since it has no other preconditions, it is active whenever the system is in this state. The effect of this process is expressed in the flow conditions for that state which show that the distance variable, $d$, changes as the value of velocity, $v$, which changes in turn as the value of acceleration, $a$. These constraints create a system of differential equations describing the simultaneous effects of velocity and acceleration on the system.

We now consider the case in which events are included but there can only be one process active on any one fluent at any one time.

In planning domains it is important to distinguish between state changes that are deliberately planned, called actions, and those, called events, that are brought about spontaneously in the world. There is no such distinction in the HA, where all control switches are called events. This distinction complicates the relationship between plans and traces, because plans contain only the control switches that correspond to actions. Any events triggered by the evolution of the domain under the influence of planned actions must be inferred and added to the sequence of actions in order to arrive at the corresponding traces.

Henzinger *et al.* (1998) discuss the use of $\epsilon$-moves, which are transitions that are not labelled with a corresponding control-switch, but with the special label $\epsilon$. The significance of these transitions is that they do not appear in traces. A trace that corresponds to an accepting run using $\epsilon$-moves will contain only those transitions that are labelled with elements from $\Sigma$. Where $\epsilon$-moves appear between time transitions, the lengths of these transitions can be accumulated into a single transition in the corresponding trace. The purpose of these silent transitions is that they allow special book-keeping transitions to be inserted into automata that can be used to simulate automata with syntactically richer constraints, allowing various reducibility results to be demonstrated. The convenient aspect of the $\epsilon$-moves is that they do not affect traces when they are transferred from the original automata to the simulations.

In PDDL+, events are similar to $\epsilon$-moves in that they do not appear explicitly in plans. However, in contrast to $\epsilon$-moves, applicable events are always *forced* to occur before any actions may be applied in a state.

When extending event-free planning instances to include events we require a mechanism for capturing the fact that events occur at the instant at which they are triggered by the world, and not at the convenience of the planner. No time must be allowed to pass between the satisfaction of event preconditions and the triggering of the event. In our semantic models we use a variable to measure the amount of time that elapses between the preconditions of an event becoming true and the event triggering. Obviously this quantity, which we call *time-slip*, must be 0 in any valid planning instance. In the HA that we construct we associate an invariant with each vertex in the control graph to enforce this requirement. It might appear that a simpler way to handle events would be to simply assert that an invariant condition for each state is that the preconditions of all events are false, while each event is represented by an outgoing transition with a jump condition specifying the precondition of the corresponding event. However, it is not possible for a jump condition on a transition to be inconsistent with the invariant of the state it leaves since both must hold simultaneously at the time at which the transition is made.

The variable used to monitor time-slip for a planning instance of dimension $n$ is the variable $X_{n+1}$. This variable operates as a clock tracking the passage of time once an event becomes applicable. We define the *time-slippage* proposition to switch the clock on whenever the preconditions of any event become true in any state. When no event is applicable the clock is switched off.

**Definition 11 Time-slippage** *For a planning instance of dimension n the variable $X_{n+1}$ is called the* time-slip *variable and* time-slippage *is defined as follows.*

$$time\text{-}slippage(\mathcal{R}) = (\dot{X}_{n+1} = 0 \lor \dot{X}_{n+1} = 1) \land (\bigvee_{e \in \mathcal{R}} \mathcal{N}(Pre_e) \to \dot{X}_{n+1} = 1)$$

*where $\mathcal{R}$ is a set of ground events.*

The use of time-slip allows us to model PDDL+ domains directly in Hybrid Automata in a standard form. An alternative would be to introduce a modified definition of Hybrid Automata that makes explicit distinction between controllable and uncontrollable transitions (actions and events respectively) and then require that uncontrollable transitions should always occur immediately when their jump conditions are satisfied. This approach would lead to an essentially equivalent formalism, but would complicate the opportunity to draw on the existing body of research into Hybrid Automata, which is why we have followed the time-slip approach.

The interpretation of a Uniprocess Planning Instance extends the interpretation of the Event-free Uniprocess Planning Instance. The added components are underlined for ease of comparison.

**Definition 12 Semantics of a Uniprocess Planning Instance** *A unary process planning instance $I = (Dom, Prob)$ is interpreted as a Hybrid Automaton, $H_I$, as follows:*

- **Variables.** *The variables of $H_I$ are $X = \{X_1, \ldots, X_{n+1}\}$, where $n$ is the dimension of the planning problem.*

  *The $n + 1th$ variable is a special control variable used to measure time-slip.*

- **Control Graph.** *The set of vertices $V$ is formed from all subsets of the ground atoms of the planning instance. The set of edges in $E$ contains an edge $e$ between $v$ and $v'$ iff there is an action or event, $a$, relevant in $v$ and:*

  $$v' = (v - Del_a) \cup Add_a$$

  *The action or event $a$ is associated with the edge $e$.*

- **Initial, invariant and flow conditions.** *The vertex labelling function* init *is defined as:*

  $$init(v) = \begin{cases} false & \text{if } v \neq Init_{logical} \\ \mathcal{N}(Init_{numeric}) \land \bigwedge \{X_i = \perp \mid X_i \notin \mathcal{N}(Init_{numeric})\} & \text{otherwise} \end{cases}$$

  *The vertex labelling function* inv *is the simple proposition that ensures time-slip is zero.*

  $$inv(v) = (X_{n+1} = 0)$$

  *The vertex labelling function* flow *is defined:*

  $$flow(v) = (\bigwedge_{i=1}^{n} \quad \{ \bigwedge_{\pi \in \mathcal{P}_v|_{X_i}} unary\text{-}context\text{-}flow_v(i, \pi) \land$$
  $$(\bigwedge_{\pi \in \mathcal{P}_v|_{X_i}} \neg \mathcal{N}(Pre_\pi) \to \dot{X}_i = 0) \} \land time\text{-}slippage(\mathcal{E}_v))$$

- **Jump conditions.** *The edge labelling function* jump *is defined as follows. Given an edge e from vertex v, associated with an action a:*

$$jump(e) = \mathcal{N}(Pre_a) \wedge UF_a(X) = X' \wedge \bigwedge_{ev \in \mathcal{E}_v} \neg\mathcal{N}(Pre_{ev})$$

  **Given an edge $e$ from vertex $v$, associated with an event $ev$:**

$$jump(e) = \mathcal{N}(Pre_{ev}) \wedge UF_{ev}(X) = X'$$

  *where $UF_a$ ($UF_{ev}$) is the updating function for action a (event ev) respectively.*

- **H-events.** *$\Sigma$ is the set of all names of all ground actions and events. The edge labelling function* hevent : $E \rightarrow \Sigma$ *assigns to each edge the name of the action or event associated with the edge.*

In this case, the flow condition says the same thing as for the event-free uniprocess planning instance, but with the additional constraint that whenever the precondition of an event is satisfied, the time-slip variable must increase at rate 1 (and may increase at rate zero otherwise). Since the invariant condition for every state insists that the time-slip variable is never greater than 0, a valid trace for this machine cannot rest in a state for any period of time once the preconditions of an event become true. It can also be seen that the jump condition for action transitions asserts that all event preconditions must be false. This ensures that events are always applied before action transitions are permitted. We now extend the preceding simple example domain to include an event, to illustrate the construction described in Definition 12:

```
(:event engineExplode
 :parameters ()
 :precondition (and (running) (>= (a) 1) (>= (v) 100))
 :effect (and (not (running)) (assign (a) 0) (engineBlown))
)
```

The corresponding machine is shown in Figure 11. The structure of this machine is similar to the previous example, but includes an extra state, reachable by an event transition. The addition of an event also requires the addition of the time-slip variable, $T$. The behaviour of this variable is controlled, in particular, by a new flow constraint in the *running* state that ensures that if the event precondition becomes true then the time-slip starts to increase as soon as any time passes. The addition of this variable and its control also propagates into the jump and flow conditions of the other states.

Finally we consider the case in which concurrent process effects occur and must be combined. This is the general case to which we refer in the rest of this paper.

In any given logical state a subset of the relevant processes will be active, according to the precise valuation of the metric fluents in the current state. The proposition *context-flow(i,$\Pi$)* asserts that the rate of change of the $i$th variable is defined by precisely those processes in $\Pi$, provided that they (and only they) are all active, and is not affected by any other process. The following definition explains how the contributions to the rate of change of a variable by several different concurrent processes are combined (see also Section 4.2). This simply involves summing the contributions that are active at an instant. Here we assume without loss of generality that contributions are *increasing* effects. Decreasing effects are handled by simply negating the contributions made by these effects.

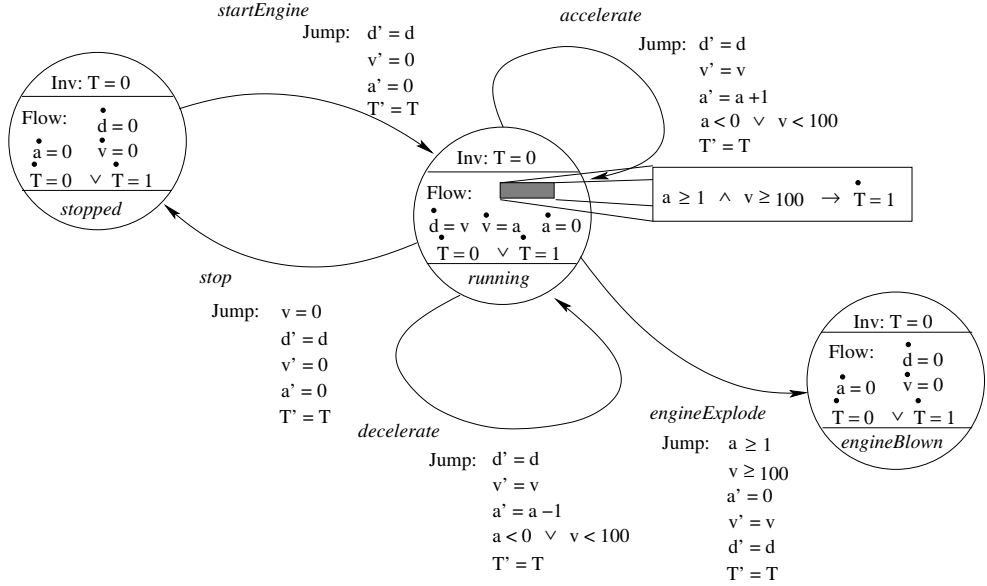

Figure 11: A hybrid automaton constructed by the translation of a uniprocess planning instance.

**Definition 13 Combined Concurrent Effects** *Given a finite set of process effects, $E = e_1 \ldots e_k$, where $e_i$ is of the form* (`increase` $P_i$ (`*` `#t` $Q_i$))*, the combined concurrent effect of $E$ on the PNE $P$, called $C(P, E)$, is defined to be*

$$\sum \{Q_i \mid i = 1, \ldots k, P = P_i\}$$

*Given a set of processes, $\Pi$, the combined concurrent effect of $\Pi$ on the PNE $P$, denoted $C(P, \Pi)$, is $C(P, E)$, where $E$ is the set of all the effects of processes in $\Pi$.*

It will be noted that if $E$ contains no processes that affect a specific variable, $P$, then $C(P, E) = 0$.

**Definition 14 Context-flow** *If $v$ is a logical state for a planning instance $I$ of dimension $n$ and $\Pi$ is a subset of $\mathcal{P}_v$, then the context-flow proposition is defined as follows.*

$$context\text{-}flow_v(i, \Pi) = (\bigwedge_{p \in \Pi} \mathcal{N}(Pre_p) \wedge \bigwedge_{p \in \mathcal{P}_v \setminus \Pi} \neg \mathcal{N}(Pre_p)) \rightarrow \dot{X}_i = C(X_i, \Pi)$$

*where $1 \leq i \leq n$.*

If $\Pi$ is empty then the context flow proposition asserts that $\dot{X}_i = 0$ for each $i$.

The interpretation of a Planning Instance extends the interpretation of a Uniprocess Planning Instance. Again, the added components are underlined for convenience.

**Definition 15 Semantics of a Planning Instance** *A planning instance $I = (Dom, Prob)$ is interpreted as a Hybrid Automaton, $H_I$, as follows:*

- **Variables.** *The variables of $H_I$ are $X = \{X_1, \ldots, X_{n+1}\}$, where $n$ is the dimension of the planning problem. The $n + 1th$ variable is a special control variable used to measure time-slip .*

- **Control Graph.** *The set of vertices $V$ is formed from all subsets of the ground atoms of the planning instance. The set of edges in $E$ contains an edge $e$ between $v$ and $v'$ iff there is an action or event, $a$, relevant in $v$ and:*

$$v' = (v - Del_a) \cup Add_a$$

  *The action or event $a$ is associated with the edge $e$.*

- **Initial, invariant and flow conditions.** *The vertex labelling function* init *is defined as:*

$$init(v) = \begin{cases} false & if\ v \neq Init_{logical} \\ \mathcal{N}(Init_{numeric}) \wedge \bigwedge \{X_i =\perp |X_i \notin \mathcal{N}(Init_{numeric})\} & otherwise \end{cases}$$

  *The vertex labelling function* inv *is the simple proposition that ensures time-slip is zero.*

$$inv(v) = (X_{n+1} = 0)$$

  *The vertex labelling function* flow *is defined:*

$$flow(v) = (\bigwedge_{i=1}^{n} \bigwedge_{\Pi \in \mathbb{P}(\mathcal{P}_v)} \underline{context\text{-}flow_v(i, \Pi)}) \wedge time\text{-}slippage(\mathcal{E}_v)$$

- **Jump conditions.** *The edge labelling function* jump *is defined as follows. Given an edge $e$ from vertex $v$, associated with an action $a$:*

$$jump(e) = \mathcal{N}(Pre_a) \wedge UF_a(X) = X' \wedge \bigwedge_{ev \in \mathcal{E}_v} \neg \mathcal{N}(Pre_{ev})$$

  *Given an edge $e$ from vertex $v$, associated with an event $ev$:*

$$jump(e) = \mathcal{N}(Pre_{ev}) \wedge UF_{ev}(X) = X'$$

- **H-events.** *$\Sigma$ is the set of all names of all ground actions and events. The edge labelling function hevent : $E \rightarrow \Sigma$ assigns to each edge the name of the action or event associated with the edge.*

The final conjunct in the jump definition for actions ensures that the state cannot be left by an action if there is an event the preconditions of which are satisfied. It is possible for more than one event to be simultaneously applicable in the same state. This is discussed further in Section 6.4.

As an illustration of this final extension in the sequence of definitions, we now add one further process to the preceding example:

```
(:process windResistance
 :parameters ()
 :precondition (and (running) (>= (v) 50))
 :effect (decrease (v) (* #t (* 0.1 (* (- (v) 50) (- (v) 50)))))
)
```

This process causes the vehicle to be slowed by a wind resistance that becomes effective from 50mph, and is proportional to the square of the speed excess over 50mph. This leads to the flow constraint in the *running* state having two new clauses that replace the original constraint on the rate of change of velocity. These new clauses are shown in the second box out on the right of Figure 12. As can be observed, the velocity of the vehicle is now governed by two different differential equations, according to whether $v < 50$ or $v \geq 50$. These equations are:

$$\frac{dv}{dt} = a, \qquad\qquad\qquad \text{if } v < 50$$
$$\frac{dv}{dt} = a - 0.1(v-50)^2, \quad \text{if } v \geq 50$$

The solution to the first is: $v = at + v_0$ where $v_0$ is the velocity at the point when the equation first applies (and $t$ is measured from this point). The solution to the second is:

$$v = \frac{c_0(\sqrt{10a} - 50)e^{-c_1 t} + 50 + \sqrt{10a}}{1 - c_0 e^{-c_1 t}}$$

where $c_1 = \frac{\sqrt{10a}}{5}$ and $c_0$ is a constant determined by the initial value of the velocity when the process first applies (and, again, $t$ is then measured from that point). As is shown by the this example, the simple differential equations that can be expressed in PDDL+ can lead to complex expressions. Of course, there is a significant difference between the provision of a semantics for this expressiveness and finding a planning algorithm that can manage it — in this paper we are concerned only with the former. We anticipate that planning will require sensible constraints on the extent to which the expressive power of PDDL+ is exploited.

The definition of the semantics of a planning instance is constructed around a basic framework of the discrete state space model of the domain. This follows a familiar discrete planning model semantics. The continuous dimensions of the model are constructed to ensure that the Hybrid Automaton will always start in an initial state consistent with the planning instance (modelled by the *init* labelling function). The state invariants ensure that no time-slip occurs in the model and that, therefore, events always occur once their preconditions are satisfied. It will be noted that since the negation of the event preconditions is added to all jump conditions for other exiting transitions, it will be impossible for a state to be exited in any other way than by the triggered event. Finally, *flow* models the effect on the real values of all of the active processes in a state. The *flow* function assigns a proposition to each state that determines a piece-wise continuous behaviour for each of the real values in the planning domain. The function is piece-wise differentiable, with only a finite number of segments within any finite interval. The reason for this is that it is possible for the behaviour of a metric fluent undergoing continuous change to affect the precondition of a process and cause the continuous change in itself or of other metric fluents to change. Such a change cannot cause discontinuity in the value of the metric fluents themselves, but it can cause discontinuity in the derivatives. The consequence of this is that the time interval between two successive actions or events might include a finite sequence of distinct periods of continuous change. As will be seen in the following section, this requires that an acceptable trace describing such behaviour will explicitly subdivide the interval into a sequence of subintervals in each of which the continuous change is governed by a stable set of differential equations.

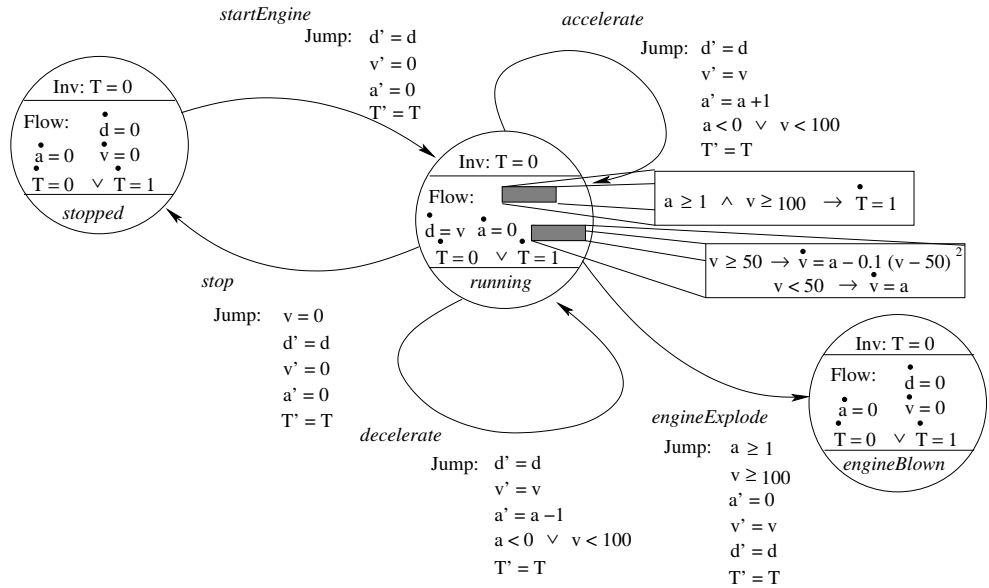

Figure 12: A hybrid automaton constructed by the translation of a planning instance.

## 6.3 Semantics of HAs

Henzinger gives a semantics for HAs by constructing a mapping to *Labelled Transition Systems* (Keller, 1976). Figure 13 shows the complete semantic relationship between planning instances and labelled transition systems and between plans and accepting runs. The top half of the figure shows the relationship already constructed by Henzinger to give a semantics to Hybrid Automata. This completes the bridge between planning instances and labelled transition systems. The details of the mapping from Hybrid Automata to labelled transition semantics are provided in this section.

The following definitions are repeated from Henzinger's paper (1996).

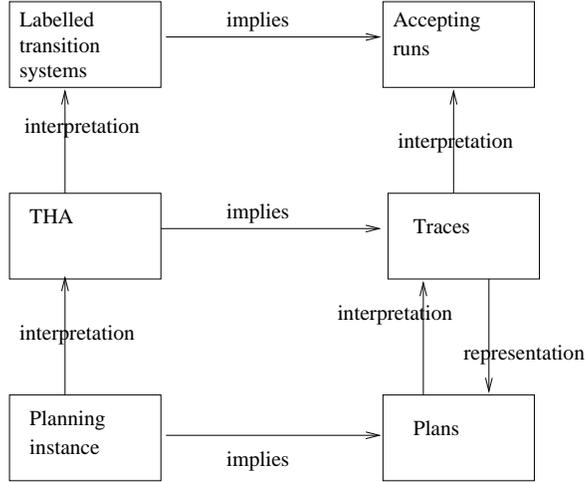

Figure 13: The semantic mapping between HAs and LTS

**Definition 16 Labelled Transition System** *A labelled transition system, S, consists of the following components:*

- **State Space.** *A (possibly infinite) set, Q, of states and a subset, $Q^0 \subseteq Q$ of initial states.*

- **Transition Relation.** *A (possibly infinite) set, A, of labels. For each label $a \in A$ a binary relation $\xrightarrow{a}$ on the state space Q. Each triple $q \xrightarrow{a} q'$ is called a transition.*

**Definition 17 The Transition Semantics of Hybrid Automata** *The timed transition system $S_H^t$ of the Hybrid Automaton H is the labelled transition system with components $Q, Q^0, A$ and $\xrightarrow{a}$, for each $a \in A$, defined as follows:*

- *Define $Q, Q^0 \subseteq V \times \mathbb{R}^n$, such that $(v, \vec{x}) \in Q$ iff the closed proposition $inv(v)[X := \vec{x}]$ is true, and $(v, \vec{x}) \in Q^0$ iff both $init(v)[X := \vec{x}]$ and $inv(v)[X := \vec{x}]$ are both true. The set Q is called the state space of H.*

- *$A = \Sigma \cup \mathbb{R}_{\geq 0}$.*

- *For each event $\sigma \in \Sigma$, define $(v, \vec{x}) \xrightarrow{\sigma} (v', \vec{x}')$ iff there is a control switch $e \in E$ such that: (1) the source of e is v and the target of e is $v'$, (2) the closed proposition $jump(e)[X, X' := \vec{x}, \vec{x}']$ is true, and (3) $hevent(e) = \sigma$.*

- *For each non-negative real $\delta \in \mathbb{R}_{\geq 0}$, define $(v, \vec{x}) \xrightarrow{\delta} (v', \vec{x}')$ iff $v = v'$ and there is a differentiable function $f : [0, \delta] \to \mathbb{R}^n$, with the first derivative $\dot{f} : (0, \delta) \to \mathbb{R}^n$ such that: (1) $f(0) = \vec{x}$ and $f(\delta) = \vec{x}'$ and (2) for all reals $\epsilon \in (0, \delta)$, both $inv(v)[X := f(\epsilon)]$ and $flow(v)[X, \dot{X} := f(\epsilon), \dot{f}(\epsilon)]$ are both true. The function f is called a witness for the transition $(v, \vec{x}) \xrightarrow{\delta} (v', \vec{x}')$.*

It is in this last definition that we see the requirement that each interval of continuous change in a timed transition system should be governed by a single set of differential equations, with a single solution as exhibited by the (continuous and differentiable) witness function.

The labelled transition system allows transitions that are arbitrary non-negative intervals of time, during which processes execute as dictated by the witness function for the corresponding period. In our definition of plans (Definition 3) we only allow rational times to be associated with actions, with the consequence that only rational intervals can elapse between actions. This means that plans are restricted to expressing only a subset of the transitions possible in the labelled transition system. We will return to discussion of this point in the following section, where we consider the relationship between plans and accepting runs explicitly.

## 6.4 Interpretation of Plans and Traces

We complete the two-layered semantics presented in Figure 13 by showing how a plan is interpreted using Henzinger's notion of trace acceptance. This will conclude our presentation of the formal semantics of PDDL+.

Using Henzinger's syntax as a semantic model, we first map plans to traces and then rely on the interpretation of traces in terms of accepting runs. A plan is a set of time-stamped actions (Definition 3): neither events nor processes appear in the specification of a plan, even though the planned actions can initiate them. By contrast, since Henzinger does not distinguish between actions and events, a trace through a HA contains control switches which might be actions or events, together with explicit time intervals between them. Plans are finite, so we are concerned only with finite traces, but plans are normally subsets of traces, because of the missing events and the possible subdivision of intervals between actions into distinct subintervals of continuous activity.

We define a new structure, the *plantrace*, which contains the sequence of control switches corresponding to the actions in a plan being interpreted. We then map plantraces to sets of traces and proceed as indicated above.

**Definition 18 Plantrace** *Let $H$ be a Hybrid Automaton, with h-event set $\Sigma$ partitioned into two subsets, $A$, the actions, and $E$, the events. A* plantrace *of $H$ is an element of the language $(\mathbb{Q}_{>0} A^+)^*$.*

A plantrace consists of sequences of one or more action control switches (denoted by $A^+$), each following a single time interval which must be greater than 0 in length (0 length intervals are not allowed because the actions on either side of them would actually be occurring simultaneously). For example, the sequence $\langle 3\, a_0\, a_1\, a_2\, 2.7\, a_3\, a_4 \rangle$ is a plantrace. Note that we have only allowed rational valued intervals between actions. This is to be consistent with the history of PDDL in which irrational time points have not been considered.

In the semantics of Hybrid Automata actions that occur at the same time point are considered to be sequenced according to the ordering in which they are recorded in the trace. In reality, it is not possible to execute actions at the same time and yet ensure that they are somehow ordered to respect the possible consequences of interaction. In order to respect this constraint we introduce an additional element in the interpretation of plans: we consider the impact of ordering actions with the same time stamp in a plan in all possible permutations in order to confirm that there are no possible interactions between them. This motivates the following definition:

**Definition 19 Permutation equivalent plantraces** *Two plantraces $\tau_1$ and $\tau_2$, are permutation equivalent, written $\tau_1 \equiv \tau_2$ if $\tau_1$ can be transformed into $\tau_2$ by permuting any subsequence that contains only actions.*

**Definition 20 Plan projection** *The projection of a plan, $P$, yields a plantrace $proj(P)$ as follows. Assume that the plan (a sequence of pairs of times and action instance names) is given sorted by time:*

$$
\begin{array}{rcl}
proj(P) & = & proj2(0, P) \\
proj2(t, \langle\rangle) & = & \langle\rangle \\
proj2(t, \langle(t_1, a), rest\rangle) & = & \langle a\rangle + proj2(t_1, rest), \ if \ t = t_1 \\
& = & \langle t_1 - t, a\rangle + proj2(t_1, rest), \ otherwise
\end{array}
$$

Plan projection is a functional description of the process by which plans are interpreted as plantraces. This process involves constructing the sequence of intervals between collections of actions that share the same time of execution, interleaved with the sequences of actions that occur together at each execution time. The most significant point to make is that where actions are given the same time for execution, the order in which they occur in the plantrace is determined simply by the (arbitrary) order in which they are listed in the plan. This does not affect the interpretation of the plan as we see in the following definition.

**Definition 21 Interpretation of a plan** *The interpretation of a plan $P$ for planning instance $I$ is the set $PT_P$ of all plantraces for $H_I$ that are permutation equivalent to $proj(P)$.*

By taking all plantraces that are permutation equivalent to the projection of the plan we remove any dependency of the interpretation of a plan on the ordering of the actions with the same time stamp. Our objective is to link the validity of plans to the acceptance of traces, so we now consider trace acceptance.

Henzinger defines trace acceptance for a trace of an HA. The definition is equivalent to the following:

**Definition 22 Trace Acceptance** *A trace, $\tau = \langle a_i \rangle_{i=1,\ldots,n}$ where $a_i \in \Sigma \cup \mathbb{R}_\geq$ is accepted by $H$ if there is a sequence $r = \langle q_i \rangle_{i=0,\ldots,n}$, where the $q_i$ are states in the timed transition system $S_H^t$ of $H$, and:*

- *$q_0 \in Q^0$.*

- *For each $i = 1, \ldots, n$, $q_{i-1} \xrightarrow{a_i} q_i$ is a transition in $S_H^t$.*

*$r$ is called an* accepting run *for $\tau$ in $H$ and, if $q_n = (v, \vec{x})$, we say that it* ends in state $v$ with final values $\vec{x}$.

A plan does not contain the transitions that represent events. For this reason, we make the following definition:

**Definition 23 Trace abstraction** *A trace, $\tau$, for a Hybrid Automaton with h-events partitioned into two sets, the actions $A$ and the events $E$, is* abstracted *to create a plantrace by removing all events in $\tau$ and then replacing each maximal contiguous sequence of numbers with a single number equal to the sum of the sequence. Finally, if the last value in this modified trace is a number it is removed.*

We can now use these definitions in the interpretation of plantraces. To avoid unnecessary multiplication of terms, we reuse the term *accepted* and rely on context to disambiguate which form of acceptance we intend in its use.

**Definition 24 Plantrace Acceptance** *Given a Hybrid Automaton, $H$, with h-event set $\Sigma$ partitioned into action $A$ and events $E$, a plantrace, $\tau$, is* accepted *by $H$ if there is a trace $\tau'$ that is accepted by $H$ such that $\tau$ is an abstraction of $\tau'$.*

It is important to observe that this definition implies that checking for acceptance of a plantrace could be computationally significantly harder than checking for standard trace acceptance. This is because the test requires the discovery of events that could complete the gaps between actions in the plantrace. However, since the events are constrained so that if they are applicable then they are forced to be applied, provided we restrict attention to commuting events, the problem of determining plantrace acceptance does not involve searching through alternative event sequences. If reasonable constraints are placed on the kinds of event cascades that may interleave between actions, the problem of checking plantrace acceptance becomes straightforward.

Finally, we return to plans and consider which plans are actually valid.

**Definition 25 Validity of a Plan** *A plan $P$, for planning instance $I$, is* valid *if all of the plantraces in $PT_P$ are accepted by the Hybrid Automaton $H_I$. The plan* achieves the goal *$G$ of $I$ if every accepted trace with an abstraction in $PT_P$ ends in a state that satisfies $G$.*

The constraint that simultaneously executed actions are non-mutex is sufficient to ensure that it is only necessary to consider one representative from the set of all permutation equivalent plantraces in order to confirm validity of a plan.

The definitions we have constructed demonstrate the relationship between plans, plantraces, traces and accepting runs. It can be observed that the definitions leave open the possibility that events will trigger in a non-deterministic way. This possibility arises when more than one event is applicable in the same state and the events do not commute. In this case, a non-deterministic choice will be made, in any accepting run, between all the applicable events. It is not possible for an action to execute before any of the events in such a state because of the time-slip process, but that process does not affect which of the events is applied. The non-deterministic choice between applicable events would allow PDDL+ to capture actions with non-deterministic outcomes. For the purposes of this paper we restrict our attention to event-deterministic planning instances.

**Definition 26 Event-deterministic Planning Instances** *A* PDDL+ *planning instance, $I$, is* event-deterministic *if in every state in $H_I$ in which two events, $e_1$ and $e_2$, are applicable, the transition sequences $e_1$ followed by $e_2$ and $e_2$ followed by $e_1$ are both valid and reach the same resulting state. In this case $e_1$ and $e_2$ are said to* commute.

If every pair of events that are ever applicable in any state commute then the planning instance is event-deterministic. In general, deciding whether a planning instance is event-deterministic is an expensive operation because the entire state space must be enumerated. However, it is much easier to construct event-deterministic planning instances because it is only necessary to consider whether pairs of events will commute. In particular, non-mutex events will always commute.

We conclude by making some observations about the relationship between plans and accepting runs. Firstly, every valid plan corresponds to a collection of accepting runs for the labelled transition system that corresponds to the HA that is the interpretation of the planning instance. The only difference between accepting runs corresponding to a given plan is in the order in which events or actions are executed at any given single time point. In contrast, there can be accepting runs for which there is no corresponding plan. This situation arises when a domain admits accepting runs with actions occurring at irrational time points. It would be possible to extend plans to allow irrational timestamps for actions. The restriction to rationals is based on the fact that an explicit report of a plan generated by a planner can only make use of timestamps with a finite representation, so there are only countably many plans that can be expressed. The fact that there are uncountably many possible transitions based on the use of arbitrary real time values is of no use to us if a planner cannot express them all. A further point of relevance is the observation, discussed in (Gupta et al., 1997), that in constructing plans for execution it is of no practical interest to rely on measurement of time to arbitrary precision. Instead, it is more appropriate to look for plans that form the core of a fuzzy tube of traces all of which are accepted. In this case, the difference between rational and irrational timestamps becomes irrelevant, since any irrational value lies arbitrarily close to a rational value, so any robust plan can be represented using rational timestamps alone. We consider that this is an important direction for future exploration, where planning problems require an additional specification of a metric and a size for the fuzzy tube that the solution plan must define, with traces in the tube having a high probability of acceptance (Fox, Howey, & Long, 2005).

## 7. Analysis

In the previous sections we have constructed a semantics for PDDL+ by mapping to Hybrid Automata and we have constructed a formal relationship between plans and traces. As a consequence, we can demonstrate that in general PDDL+ domains provide succinct encodings of their corresponding Hybrid Automata, since a PDDL+ model can have a state space that is exponential in the size of the encoding.

Having established the relationship between PDDL+ domains and Hybrid Automata, we can now benefit from the large body of research into Hybrid Automata and their sub-classes. One issue that has been widely addressed is the boundary between decidable and undecidable classes of Hybrid Automata and this boundary can be reinterpreted for the reachability question for interesting subsets of the PDDL+ language.

In the following we only consider subsets of PDDL+ that are interesting in the sense of modelling different kinds of restricted continuous temporal behaviours.

### 7.1 Reachability within Hybrid Automata

The Reachability problem for Hybrid Automata is the problem of determining, given an automaton $H$, whether there is a trajectory of the timed transition system, $S_H^t$, that visits a state of the form $(v, \mathbf{x})$. In the context of Hybrid Automata, $v$ is typically an error state, so that Reachability questions are posed to determine whether an automaton is safe. In the context of planning, the Reachability question is equivalent to Plan Existence. As is discussed in Henzinger's paper (1996), the general Reachability question for Hybrid Au-tomata is undecidable. This is unsurprising, since introducing metric fluents with arbitrary behaviours into the language results in sufficient expressive power to model Turing Machine computations. Indeed, Helmert has shown (2002) that even relatively simple operations on discrete metric variables are sufficient to create undecidable planning problems. However, under various constraints, Reachability is decidable for several kinds of hybrid system, including *Initialised Rectangular Automata* (Henzinger et al., 1998). In the following discussion we restrict our attention to deterministic Initialised Rectangular Automata, focussing particularly on Timed Automata (Alur & Dill, 1994), Priced Timed Automata (Rasmussen et al., 2004) and Initialised Singular Automata (Henzinger, 1996) and their relationships with fragments of the PDDL family of languages.

To simplify the definition of Rectangular Automata we introduce the following definition:

**Definition 27 Interval Constraint** *Any constraint on a variable $x$ of the form $x \bowtie c$ for a rational constant $c$ and $\bowtie \in \{\leq, <, =, >, \geq\}$ is an* interval constraint. *A conjunction of interval constraints is also an interval constraint.*

Rectangular Automata are Hybrid Automata in which all initial, invariant and flow conditions are interval constraints, whose flow conditions refer only to variables in $\dot{X}$ and whose jump conditions are a conjunction of interval constraints and constraints of the form $x'_i = x_i$. Rectangular Automata may be non-deterministic because the interval constraints that determine the initial, flow and jump conditions might not determine unique values for the variables they constrain. Initialised Rectangular Automata meet the additional constraint that for each control switch, $e$, from $v$ to $w$, if $flow(v)_i \neq flow(w)_i$ then $x'_i = x_i$ does not appear in $jump(e)$.

Initialised Rectangular Automata are important because they represent the boundary of decidability for Hybrid Automata (Henzinger et al., 1998).

Since PDDL is a deterministic language, we are most interested in the deterministic version of Rectangular Automata. A *Singular Automaton* is a Rectangular Automaton with deterministic jumps and variables of finite slope (that is, the flow conditions determine a unique constant rate of change for each variable). It is *initialised* if each continuous variable is reset every time its rate of change is altered by the flow function. The fact that the variables have finite slope allows rates of change to be modified. Initialised Singular Automata allow the modelling of linear continuous change and the rate of change to be modified provided that the corresponding variable is reset when this occurs. This constraint prevents the modelling of *stopwatches* (Alur & Dill, 1994), so ensures decidability of the Reachability problem. Whilst quite expressive, such automata cannot capture the dynamics that arise in many continuous planning domains. For example, it would not be possible to model the effect, on the *level* of water in a tank, of adding a second water supply some time into the filling process. According to the reset constraint the *level* value would have to be reset to zero when the second water source is introduced.

A Timed Automaton is a Singular Automaton in which every variable is a clock (a one-slope variable). These automata can be used to model timed behaviour and to express constraints on the temporal separation of events (Alur & Dill, 1994). However, they cannot be used to model the continuous change of other quantities because clocks cannot be used to store intermediate values. They can be stopped and reset to zero, but they cannot operate as memory cells. Alur and Dill (1994) prove that reachability is decidable in Timed Automata because the infinite part of the model (the behaviour of the clocks) can be decomposed into a finite number of *regions*. This demonstrates that the infinite character of Timed Automata can be characterised by an underlying finite behaviour. Helmert (2002) used a similar regionalisation technique in proving that the plan existence question for planning models including certain combinations of metric conditions and effects remains decidable.

A Priced Timed Automaton is a Timed Automaton in which costs are associated with the edges and locations of the automaton. Costs are accumulated over traces and enable a preference ordering over traces. PTAs have been used to solve simple Linear Programming and scheduling problems (Rasmussen et al., 2004). Costs behave like clocks except that they can be stopped, or their rates changed, without being reset. To retain decidability despite the addition of cost variables their use is restricted: costs cannot be referred to in jump conditions although they can be updated following both edge and delay transitions. PTAs have been used to model planning problems in which actions are associated with linearly changing costs. For example, the *Airplane* scheduling bench mark, in which planes incur cost penalties for late and early landing, can be expressed in the syntax of PDDL+ and solved using PTA solution techniques (Rasmussen et al., 2004). The models are very restrictive: the dependence of the logical dynamics of the domain on the cost values cannot be expressed so costs cannot, for example, be used to keep track of resource levels during planning when resources are over-subscribed. However, they are capable of expressing a class of problems which require the modelling of continuous change and can therefore be seen as a fundamental step towards the modelling of mixed discrete-continuous domains.

Since PTAs allow the modelling of non-trivial planning domains with continuous change we begin by constructing a fragment of PDDL+ that yields state-space models exactly equivalent to PTAs. In the remainder of this section we then discuss the relationships between richer PDDL+ fragments and different automata.

## 7.2 PDDL+ and Priced Timed Automata

Rasmussen *et al.* (2004) describe the components of a Priced Timed Automaton (PTA) as follows. It contains a set of real-valued clocks, $C$, over which constraints can be expressed by a set of clock constraints, $B(C)$. There is a set of actions, $Act$. A PTA is given as a 5-tuple $(L, l_0, E, I, P)$, where L is a finite set of locations (that is, states), $l_0$ is the initial location, $E \in L \times B(C) \times Act \times 2^C \times L$, $I : L \rightarrow B(C)$ is a function assigning invariants to locations and $P : (L \cup E) \rightarrow N$ assigns prices to edges and locations. An edge, $E = (l, b, a, r, l')$ is a transition between locations $l$ and $l'$ using action $a$, which has the jump condition $b$ (over the clocks) and resets the clocks in $r$ to zero. The price function represents a discrete cost for transitions and a continuous cost associated with staying in each location.

For the modelling of arbitrary Priced Timed Automata we define a PDDL+ fragment which we refer to as PDDL+$_{PTA}$. The following definition of this fragment is not unique. Other subsets of PDDL+ exist with the same expressive power, including subsets directly relying on processes and events. By excluding events in this fragment we are trivially guaranteed to have a deterministic language.

The PDDL+$_{PTA}$ fragment uses the standard PDDL+ language features, but subject to the following constraints:

1. Processes may have only logical preconditions and exactly one process must be active in each state except, possibly, one special state, *error*.

2. Each process must increase all the metric fluents at rate 1, except for one special variable, $c$, which may be increased at any constant rate.

3. Each action may have preconditions that refer to any of the metric fluents except $c$. These preconditions must appear in any of the forms (called *clock constraints*) conforming to the following: $(\bowtie \ x_i \ n)$ or $(\bowtie \ (\texttt{-} \ x_i \ x_j) \ m)$ where $n, m$ are natural numbers and $\bowtie \in \{<, \leq, =, \geq, >\}$.

4. Each action may reset the value of any metric fluent, other than $c$, to zero. They may increase the value of $c$ by any constant value.

5. The domain may contain events whose precondition may include literals and clock constraints. The effect of every event must be to leave the system in the special state *error*, from which no further transitions are possible.

6. The plan metric is of the form: `(:metric minimize (c))`.

These constraints ensure that the state space yielded by a PDDL+$_{PTA}$ description, and its transition behaviour, is equivalent (within a constant multiple encoding size) to a PTA. We cannot demonstrate direct equivalence between PDDL+$_{PTA}$ and PTAs because PDDL+$_{PTA}$ models can be exponentially more compact than the explicit PTA to which it corresponds. This compaction is similar to that which can be obtained by factorising PTA models (Dierks, 2005). We demonstrate the indirect equivalence of the two languages in Theorem 2.

**Definition 28 Indirectly Equivalent Expressiveness** *Given two languages, $L_1$ and $L_2$, $L_1$ is indirectly equivalent in expressive power to $L_2$ if each sentence, $s_1$, in $L_1$ defines a model, $M_{s_1}$, such that $L_2$ can express $M_{s_1}$ with only a polynomial increase in encoding size (in the size of $M_{s_1}$) and each sentence, $s_2$, of $L_2$ can be expressed in $L_1$ with at most a polynomial increase in the size of the encoding (in the size $s_2$).*

We note that this definition is asymmetric: sentences of $L_1$ define models that can be expressed efficiently in $L_2$, but sentences of $L_2$ can be efficiently expressed directly in $L_1$. The sentences of $L_1$ might be compact encodings of their corresponding models and therefore we cannot claim a direct equivalence in expressiveness between $L_1$ and $L_2$. This is intentional and allows us to exploit the PDDL property that it is a compact encoding language for the state spaces and transition behaviours of the planning domains it defines.

**Theorem 2** *PDDL+$_{PTA}$ has indirectly equivalent expressive power to Priced Timed Automata.*

The proof of this theorem can be found in Appendix B.

**Lemma 1** *A sentence, $s$, of PDDL+ defines a transition system that is at most doubly exponential in the size of $s$.*

This is straightforward: the set of literals defined by a PDDL+ problem instance is at most exponential in the number of parameters in the predicate of the highest arity and the state space this defines is at most exponential in the size of the set of literals.

<div align="right">□</div>

**Corollary 1** *Reachability in* PDDL+$_{PTA}$ *is decidable.*

This follows from Theorem 2, Lemma 1 and the decidability of PTAs (Larsen, Behrmann, Brinksma, Fehnker, Hune, Pettersson, & Romijn, 2001).

<div align="right">□</div>

PDDL+$_{PTA}$ can be extended, while remaining decidable, by allowing additional metric fluents whose (non-continuous) behaviour is constrained according to one of the decidable subsets identified by Helmert (2002), provided that these fluents are distinct from the clock and cost variables.

We can similarly define a language fragment with expressive power equivalent to Initialised Singular Automata. Initialised Singular Automata represent the most expressive form of deterministic automaton with a decidable Reachability problem. Their constraints limit the extent to which we can reason about the dependence of behaviour on temporal and metric quantities.

In an Initialised Singular Automaton continuous change is not restricted to variables of slope 1, in contrast to the clocks of Timed Automata. All variables have finite slope (that is, their rates of change can only take one of a finite set of different values), but must be initialised to a given constant whenever their rates of change are altered. The values of these variables can be referred to in the jump conditions of the automaton.

For the modelling of Initialised Singular Automata we define a PDDL+ fragment which we refer to as PDDL+$_{ISA}$. This fragment contains the same syntactic components as PDDL+$_{PTA}$ but is subject to slightly different constraints:

1. Processes may have only logical preconditions and exactly one process must be active in each state except, possibly, one special state, *error*.

2. Each process must increase all the metric fluents at constant rates.

3. Each action may have preconditions that refer to any of the metric fluents. These preconditions must appear in any of the forms (called *clock constraints*): $(\bowtie\ x_i\ n)$ or $(\bowtie\ (-\ x_i\ x_j)\ m)$ where $n, m$ are natural numbers and $\bowtie\ \in\ \{<, \leq, =, \geq, >\}$.

4. Each action may reset the value of any metric fluent to any constant natural number value. Any action that causes a transition to a state where a new process is active must also reset all metric fluents whose rates of change are different under the new process.

5. The domain may contain events whose precondition may include literals and clock constraints. The effect of every event must be to leave the system in the special state *error*, from which no further transitions are possible.

This fragment does not contain a special plan metric: problems in PDDL+$_{ISA}$ are of interest for the plan existence problem.

**Theorem 3** PDDL+$_{ISA}$ *has indirectly equivalent expressive power to Initialised Singular Automata.*

The proof of this result is analogous to Theorem 2. The constraints ensure that the behaviour of Initialised Singular Automata is captured in this language and that the language contains only expressions that can be effectively modelled within Initialised Singular Automata.

It is of interest to review the domains we have considered in this paper, the planetary lander domain and the accelerating vehicle domain, with respect to the modelling power of PDDL+$_{PTA}$ and PDDL+$_{ISA}$. In the first instance, since both domains involve non-linear change, in the power generation and state of charge curves and in the distance travelled by the vehicle when acceleration is non-zero, it is clear that both domains lie outside the modelling power of these constrained languages. It is feasible to consider approximating the non-linear behaviour in some cases. For example, the generation curve might be approximated by a small set of linear functions (with a corresponding loss of opportunities to exploit the margins of the model). The generation curve can be tied to absolute points on the timeline and the values of the points at which the linear functions would be required to meet in order to approximate the original curve can be identified in advance. In order to create a reasonable approximation, the functions would require different slopes (shallow at the start, then steeper and then shallower again, to approximate the first half of the bell curve), which cannot be achieved using PDDL+$_{PTA}$, since this only allows clock variables. PDDL+$_{ISA}$ is powerful enough to express differently sloped linearly changing variables of this kind. Unfortunately, the state of charge curve, even with approximations, is beyond the expressive power of either language. For PDDL+$_{ISA}$ the problem is that there must be a memory of the old state of charge when the slope of the charge curve changes. With PDDL+$_{PTA}$ the cost variable could be used to model the state of charge, since it has the capacity to both be used as a memory and to have different rates of change. However, the cost variable cannot be used in preconditions of any actions, which means that any attempt to model the battery in this way would force a decoupling between the battery state of charge and the actions that use power. A more realistic model of the battery management lies outside the power of PDDL+$_{PTA}$ and PDDL+$_{ISA}$.

However, other interesting problems involving continuous change are perfectly amenable, as Dierks (2005) and Behrmann *et al.* (2005) have shown. The Aircraft Landing problem (Beasley, Krishnamoorthy, Sharaiha, & Abramson, 2000) can be modelled as a PTA because it has one source of continuous change which can be modelled as a cost variable. In this domain, some number of aircraft must land on a single airport, sometime between an earliest and latest landing time and as close to a target time as possible. The earliest, latest and target times are defined for each aircraft. A cost is associated with the problem and there is a charge associated with landing a plane early or late, where the charge decreases linearly from the earliest landing time towards the target time and increases linearly from the target up to the latest landing time. The behaviour of the aircraft is not dependent on the value of the cost variable, although the quality of a landing schedule is determined by it. The Aircraft Landing problem is an optimisation problem where the quantity being optimized changes continuously over time.

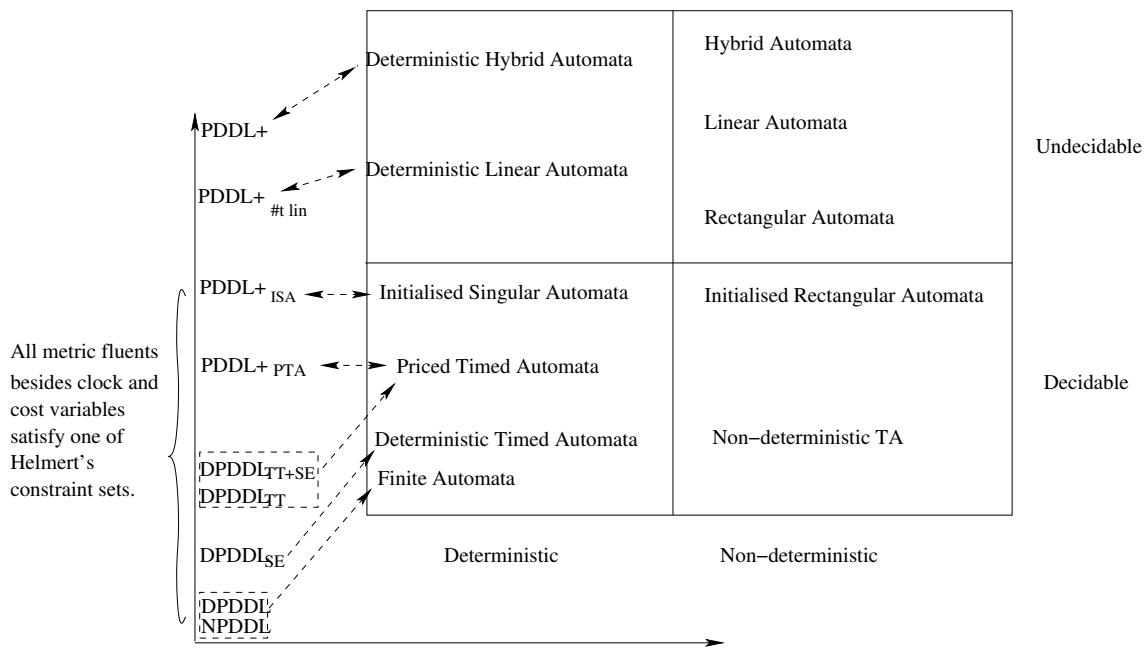

Figure 14: Mapping PDDL fragments to corresponding deterministic automata. PDDL is a family of deterministic languages. However, extension to support temporal, numeric or logical non-determinism would give access to the automata on the right hand side of the figure.

In Figure 14 we present some formal relationships between PDDL+ fragments and classes of hybrid automata. The bottom half of the figure concerns decidable fragments of PDDL+ whilst the top half concerns undecidable fragments. In the figure, DPDDL is the PDDL+ fragment in which fixed duration durative actions are allowed (and no events or processes), restricted to those with no *at start* effects and NPDDL (non-temporal PDDL) is the fragment in which they do not occur. Initial effects are significant because it is these that make it possible to create domains in which there are problems that can only be solved by exploiting concurrency. The reason for this is that if all effects are restricted to the end points of actions, then it is always possible to find a sequential plan to solve a problem for which there is a concurrent solution. In cases where concurrency is not required to solve a problem, durative actions can be sequentialised and their durations simply summed as discrete end effects. When actions can have initial effects then it is possible for there to be effects that are added at the start of an action and deleted at the end, creating windows of opportunity into which other actions must be fitted concurrently if they are to exploit the effects. Once concurrency matters it is necessary to monitor the passage of time and this requires the power of timed automata. The other language variants we show are as follows:

- DPDDL$_{SE}$ which allows start effects, raising the need for concurrency as explained above.

- DPDDL$_{TT}$, which is DPDDL together with the use of plan metrics using the term `total-time`: this also requires concurrency since the time required by a plan depends on the extent to which non-interfering actions can be selected to reduce the make-span of the plan.

- DPDDL$_{TT+SE}$ which is DPDDL together with `total-time` and allowing the use of start effects.

- PDDL+$_{\#t\,lin}$ is PDDL+ restricted to linear rates of change on all metric fluents. This restriction is insufficient to ensure decidability, but the use of linear rates of change makes it possible to apply linear constraint solvers to the problems. As a consequence, this subset of PDDL+ captures the continuous planning problems to which planning technology has already been applied (Shin & Davis, 2005; McDermott, 2003b; Wolfman & Weld, 1999; Penberthy & Weld, 1994) (in the first of these, in particular, the models are actually expressed in PDDL+).

In the bottom half of the figure both NPDDL and DPDDL include metric conditions and effects constrained to occur in decidable combinations as defined by Helmert (2002). Helmert demonstrated that adding (discrete) metric effects and conditions to the propositional PDDL fragment already adds dramatically to the expressive power of the language. In particular, with a quite limited set of (discrete) metric effects and conditions decidability is already lost. However, there are restrictions to the use of metric variables that leave a decidable fragment. In order for the extensions we are discussing to retain decidability we must not, of course, sacrifice it by adopting too rich a set of discrete metric effects or preconditions. This is why in the lower half of the figure we work within these constraints. Our results extend Helmert's by introducing continuous metric change, while demonstrating boundaries that retain decidability. In the top half of the figure these constraints are lifted. The left-hand half of the figure concerns deterministic models: PDDL+ is a language for expressing deterministic domains, so we have restricted our attention to this side. The right-hand half concerns non-deterministic variants of the models considered. We include these only to provide a general context for our results.

## 8. Related Work

In this section we discuss the relationship between PDDL+ and several other related formalisms in the literature. In addition, we consider the extent to which PDDL+ addresses the inadequacies of PDDL2.1 in terms of expressive power and convenience. Our paper on PDDL2.1 was published in the Journal of Artificial Intelligence Research Special Issue on the 3rd IPC (Fox & Long, 2003). The editors of this special issue invited five influential members of the planning community to contribute short commentaries on the language, indicating their support for, and objections to, the choices we made in the modelling of time and time-dependent change. These were Fahiem Bacchus, Mark Boddy, Hector Geffner, Drew McDermott and David Smith. We now discuss the parts of their commentaries that are relevant to the modelling of durative behaviour and continuous change and explain how we believe PDDL+ addresses the issues raised. It is interesting to note that many of the objections raised by these commentaries are addressed by the start-process-stop model of PDDL+. We begin by considering these commentaries and then go on to discuss related formalisms.

## 8.1 PDDL+ versus PDDL2.1

The commentators were invited to comment on the decisions made in PDDL2.1, and the limitations they impose on temporal domain modelling. Some of the issues raised by the commentators have been addressed in PDDL+. In this section we identify the issues that were raised that are relevant to the development of PDDL+ and explain how we think PDDL+ resolves them.

Bacchus (2003) proposes an alternative to the continuous durative actions of PDDL2.1 which has similarities with the start-process-stop model of PDDL+. Both approaches recognise that durative activity is sometimes best modelled using an underlying, interruptible process. Whilst Bacchus proposes that the initiation and running of the processes be wrapped up into durative actions with conditional effects, PDDL+ achieves the same effect with cleanly separated continuous autonomous processes and events. PDDL+ can express behaviours that are dependent on continuous variables other than time, while Bacchus' proposal is limited to purely time-dependent processes (he does not allow interacting processes so does not consider other forms of continuous change).

McDermott and Boddy have consistently supported the use of autonomous processes in the representation of continuous change. In his commentary McDermott (2003a) identifies the weaknesses of the durative action-based representation of change and argues that the continuous durative actions of PDDL2.1, which allow the modelling of duration inequalities and time-dependent effects, are "*headed for extinction in favour of straightforward autonomous processes*". The start-process-stop model of PDDL+ replaces the continuous durative actions of PDDL2.1 with constructs that fully exploit autonomous processes to support richer and more natural models of continuous domains. As we have shown in this paper, the modelling of events adds expressive power as Boddy anticipates (Boddy, 2003).

In his commentary Smith (2003) raises some philosophical objections to the durative action model of PDDL2.1 which he argues is too restrictive to support convenient models of interesting durative behaviours. In PDDL2.1 actions specify effects only at their start and end points, although conditions can be required to remain invariant over the whole durative interval. Smith proposes a durative action model that is richer than that proposed in PDDL2.1, in which effects can occur at arbitrary points within the durative interval and conditions might also be required to hold at identified timepoints other than at the start or end of an action. Although, in principle, it is possible to decompose PDDL2.1 durative actions into sequences of actions that achieve these effects, Smith correctly observes that this would generally result in inconvenient and impractical models. He argues that action representations should encapsulate the many consequences of their application in a way that frees a planner from the burden of reasoning about them in their minutiae. He observes that the computational effort involved, in stringing together the sub-actions that are required to realise a complex activity, would normally be prohibitive.

We agree that the PDDL2.1 durative action model is restrictive in forcing effects to occur only at the end points of the actions. Smith's rich durative model can be seen as encapsulating the effects of starting and ending one or more processes, together with the effects of these processes, into an action-based representation. By committing the activity to a certain amount of time these actions abstract out the time-dependent details and avoid the need for the planner to reason about them or their interactions. This is a simplification that is no doubt sufficient in many practical contexts (and indeed, is sufficient for the satellite domain that he discusses in his commentary). It might indeed be of interest to provide such representations as abstractions of the start-process-stop model.

This point goes to the heart of the contribution of this paper: we have provided a set of primitives for building modelling constructs. In providing a formal semantics for these primitives we have provided a way of interpreting abstract constructs built from these primitives. We argue that, in combination with natural modelling concepts like fixed-length durative actions, the start-process-stop primitives provide a usable planning domain description language. However, we are concerned here with the formal underpinnings of the language rather than with the modelling convenience it provides. We agree with Smith that abstract modelling constructs, built from the primitives, might enhance the modelling experience in the same way that abstract programming constructs enhance the programming experience over programming at the machine code level.

## 8.2 Related Formalisms

A number of representational formalisms have been proposed for expressing temporal and metric activity in planning. The closest recent counterpart to PDDL+ is the modelling language, Opt, of the Optop planner (McDermott, 2004). This language was developed independently, by Drew McDermott, at the same time as PDDL+ was first proposed, and some of the similarities between Opt and PDDL+ are due to discussions between the developers of the two languages at the time. Opt is a PDDL-like dialect which was strongly influenced by the work of McDermott and other authors on the PDDL family of languages. Opt supports autonomous processes which run when their preconditions are satisfied and are not under the control of the planner. Unlike PDDL+, Opt does not contain explicit events - these are embedded inside processes which run for as long as their preconditions remain true. Opt also retains durative actions as an alternative to the explicit modelling of continuous change and models timed initial literals and derived predicates. A planner, Optop, was developed by McDermott (McDermott, 2005) for the subset of Opt that models linear continuous change. Planners already exist for handling interesting subsets of PDDL+, both directly (Shin & Davis, 2005) and indirectly (Dierks, 2005). In the later case, the language of the PTA solver UPPAAL-cora (Behrmann et al., 2005) is used, which has modelling power equivalent to that of PDDL+$_{PTA}$.

The semantics of Opt processes is given in terms of infinitely many situations occurring within a finite time, each associated with different fluent values of the continuously changing variables. Opt and PDDL+ are fundamentally related to Reiter's work on continuous dynamics in the situation calculus (Reiter, 2001). McDermott developed a situation calculus semantics for Opt, whereas we have constructed an explicit relationship between PDDL+ and the theory of Hybrid Automata in order to make explicit the relationship be-tween PDDL+ planning and control theory. A similar relationship is drawn in the CIRCA architecture (Musliner, Durfee, & Shin, 1993), which integrates planning with real time control using probabilistic timed automata.

The qualitative reasoning community have proposed hybrid state-based models of dynamic physical systems (Forbus, 1984; de Kleer & Brown, 1984; Kuipers, 1984). Kuipers (1984) considers the qualitative simulation of physical systems described in terms of continuously varying parameters. He proposes a qualitative representation of the differential equations governing the behaviour of a system, expressed as systems of constraints over the key parameters describing the state of the system at discrete points in time. This representation supports commonsense reasoning about the evolution of physical systems about which quantitative reasoning would be computationally prohibitive.

Other formalisms have been developed within the fields of planning and reasoning about action and change. Temporal and resource management are provided in HSTS and Europa (Frank & Jónsson, 2003; Jónsson, Morris, Muscettola, Rajan, & Smith, 2000), Ix-TeT (Laborie & Ghallab, 1995), CIRCA (Musliner et al., 1993), LPSAT (Wolfman & Weld, 1999), Zeno (Penberthy & Weld, 1994) and HAO* (Benazera, Brafman, Meuleau, Mausam, & Hansen, 2005) to mention only a few such systems in the planning literature. For the most part, these systems are plan-generation systems using representation languages that support the restricted modelling of continuous change and metric time. By contrast, PDDL+ proposes an unrestricted representation language, and its semantics, without describing specific search algorithms for the construction of plans. Finding efficient algorithms for reasoning with PDDL+ domains is a separate topic, in which there has already been progress (Shin & Davis, 2005; Dierks, 2005; McDermott, 2004) as mentioned above. Of course, the demands of practical planning restrict how ambitious one can be in using PDDL+ to model real planning applications, but this is not a reason to impose artificial restrictions on the expressiveness of the language.

The key objective of HSTS (Heuristic Scheduling Testbed System) (Muscettola, 1993) is to maintain as much flexibility as possible in the development of a plan, so that the plan can be robustly executed in the face of unexpected events in the environment. HSTS embodies the close integration of planning and scheduling, enabling the representation of complex resource-intensive planning problems as *dynamic* constraint satisfaction problems (DCSP). In DDL, the Domain Description Language of HSTS, A distinction is made between domain *attributes* (ie: the components of the domain that can exhibit behaviours) and their states, with the activity of each attribute represented on a separate time line. No distinction is made between states and actions: actions are added to the time lines by inserting *tokens* representing predicates holding over flexible intervals of time. Each token is associated with a set of *compatibilities* which explain how they are constrained with respect to activities on the same and other time lines. Compatibilities express relations very similar to the TIMELOGIC constructs of Allen and Koomen (Allen & Koomen, 1983). Choices in the development of the plan are explored through a heuristic search and inconsistencies in the DCSPs representing the corresponding partial plans result in pruning.

Events of the kind provided by PDDL+ are expressed as disjuncts in the compatibility constraints associated with the actions that produce them. For example, the action of opening a water source to fill a tank would be expressed as a token constrained to *meet* either the event of flooding or an interval in which the water source is closed. Then, whether the tank floods or not depends on how long the interval of filling lasts, and the flooding event can be avoided by expressing the constraint that it end before the water level exceeds the tank capacity. The process by which the water level increases while the tank is filling is expressed using *sequence compatibilities*, which allow variables to take arbitrarily many contiguous values from a sequence during an interval. The actual water level at the end of the interval can be identified, using linear programming techniques, as one of the values in this sequence. The notion of *procedural reasoning* (Frank, Jönsson, & Morris, 2000) has been introduced into the framework to support efficient reasoning with rates of change on continuous variables and their interactions within a plan.

HSTS therefore supports the representation of the interaction between actions, processes and events and their exploitation in the development of flexible plan/schedules. In this respect, DDL is somewhat more expressive than PDDL+ because of the flexibility in its temporal database. Allowing intervals to last any amount of time between a specified lower and upper bound introduces bounded temporal flexibility into the reasoning framework.

IxTeT (Laborie & Ghallab, 1995) is a partial order causal link planner that uses a task representation very similar to that of discrete durative actions in PDDL2.1. A key difference is that PDDL2.1 discrete durative actions are restricted to the representation of step function change at the start and end points of the interval, whilst IxTeT tasks can have effects at any specified point during the interval. This allows piecewise continuous change to be represented. Continuous change cannot be modelled (except by means of very small intervals in the piecewise representation of the appropriate function). Furthermore, the durative actions of IxTeT are not of fixed duration — they can endure any amount of time within a specified interval. Thus, IxTeT also models bounded temporal flexibility and is able to construct *flexible* plans. IxTeT continues the traditions of POCL planning (McAllester & Rosenblitt, 1991; Penberthy & Weld, 1992) in which a plan is built up as a partially ordered graph of activities with complex flexible temporal constraints. A Simple Temporal Network (Dechter, Meiri, & Pearl, 1991) is used to determine the consistency of a given temporal constraint set. STNs, also used in HSTS, are a powerful technique for temporal reasoning but are restricted to reasoning about discrete time points.

There is an important difference between *modelling* the continuous process of change and *computing* the values of continuous-valued variables during planning. In some systems, continuous change is explicitly modelled, so that trajectories can be constructed for the variables throughout the timeline of a plan. In other systems, the continuous processes that determine the behaviour of metric variables are implicit, but the values of these variables are available, through some computation, at certain times along their trajectories.

Zeno (Penberthy & Weld, 1994) uses an explicit representation of processes as differential equations and solves them to determine whether the temporal and metric constraints of a problem are met in any partial plan (and to identify the values of continuous-valued resources when action preconditions require them). LPSAT (Wolfman & Weld, 1999) also uses an explicit model of the processes that govern continuous change (although its use of linear constraint solving limits it to linear processes). A different approach, in which the processes are not explicitly modelled, can be seen in HAO* (Benazera et al., 2005). Here, although continuous-valued resources are modelled, the way in which they change over time is not. The different possible metric outcomes of an action are discretised and associated with probabilities. Plan construction can be seen in terms of policy construction within a hybrid MDP framework. Time can be managed in the same way as any other metric resource (a certain action will take $t$ time units with probability $1-p$) so there is no need to model the passage of time directly. Whilst the time-dependent nature of the metric effects of actions can be captured in this way, actions cannot interact with the passage of time in order to exploit or control episodes of continuous change.

## 9. Conclusions

In this paper we have presented a planning domain description language, PDDL+, that supports the modelling of continuous dynamics. We have provided a formal semantics by specifying a mapping from PDDL+ constructs to those of deterministic hybrid automata. We have also related fragments of PDDL+ to automata with different levels of expressive power. Our goal has been to develop a PDDL extension that properly models the passage of time, and the continuous change of quantities over time, to support the modelling of mixed discrete-continuous planning domains.

Our primary goal has been to establish a baseline for mixed discrete-continuous modelling, and to provide a formal semantics for the resulting language. Additionally we wanted to make a strong connection between planning and automata theory in order to facilitate the cross-fertilisation of ideas between the planning and real-time systems and model-checking communities. We have explored the relationship between fragments of PDDL+ and automata-theoretic models on the boundary of decidability in order to better understand what is gained or lost in expressive power by the addition or removal of modelling constructs.

We have not focussed on how to make PDDL+ convenient for modelling because modelling convenience is not related to expressive power. We agree that it might be desirable to build abstract modelling constructs on top of the baseline language in order to enhance modelling convenience. Nevertheless, so that PDDL+ can be used for experimental purposes, in the development of technology for mixed discrete-continuous planning, we have presented it as a usable language that builds directly upon the current standard modelling language for temporal domains. We have presented two examples of domain models that exploit processes, events and durative actions in the succinct representation of continuous change. Future work will consider what more powerful modelling constructs might be built to support the convenient modelling of larger scale mixed discrete-continuous domains.

## Acknowledgments

We would like to extend special thanks to David Smith for his detailed critical analysis of earlier drafts of this paper and his many insightful comments and suggestions. We would also like to thank Subbarao Kambhampati and the anonymous referees for helping us to organise and clarify the presentation of this work, and Jeremy Frank, Stefan Edelkamp, Nicola Muscettola, Drew McDermott, Brian Williams and Mark Boddy, all of whom helped us to refine and sharpen our ideas and formulations.

## Appendix A. PDDL2.1 Core Definitions

In this appendix we present the definitions from (Fox & Long, 2003) that are most relevant to this paper, for ease of reference. For detailed discussions of these definitions and their relationships see the original source.

**Core Definition 1 Simple Planning Instance** *A simple planning instance is defined to be a pair*

$$I = (Dom, Prob)$$

*where $Dom = (Fs, Rs, As, arity)$ is a 4-tuple consisting of (finite sets of) function symbols, relation symbols, actions (non-durative), and a function arity mapping all of these symbols to their respective arities. $Prob = (Os, Init, G)$ is a triple consisting of the objects in the domain, the initial state specification and the goal state specification.*

*The primitive numeric expressions of a planning instance, $PNEs$, are the terms constructed from the function symbols of the domain applied to (an appropriate number of) objects drawn from $Os$. The dimension of the planning instance, $dim$, is the number of distinct primitive numeric expressions that can be constructed in the instance.*

*The atoms of the planning instance, $Atms$, are the (finitely many) expressions formed by applying the relation symbols in $Rs$ to the objects in $Os$ (respecting arities).*

*$Init$ consists of two parts: $Init_{logical}$ is a set of literals formed from the atoms in $Atms$. $Init_{numeric}$ is a set of propositions asserting the initial values of a subset of the primitive numeric expressions of the domain. These assertions each assign to a single primitive numeric expression a constant real value. The goal condition is a proposition that can include both atoms formed from the relation symbols and objects of the planning instance and numeric propositions between primitive numeric expressions and numbers.*

*$As$ is a collection of action schemas (non-durative actions) each expressed in the syntax of PDDL. The primitive numeric expression schemas and atom schemas used in these action schemas are formed from the function symbols and relation symbols (used with appropriate arities) defined in the domain applied to objects in $Os$ and the schema variables.*

**Core Definition 2 Logical States and States** *Given the finite collection of atoms for a planning instance $I$, $Atms_I$, a logical state is a subset of $Atms_I$. For a planning instance with dimension $dim$, a state is a tuple in $(\mathbb{R}, \mathbb{P}(Atms_I), \mathbb{R}_\perp^{dim})$ where $\mathbb{R}_\perp = \mathbb{R} \cup \{\perp\}$ and $\perp$ denotes the undefined value. The first value is the time of the state, the second is the logical state and the third value is the vector of the dim values of the dim primitive numeric expressions in the planning instance.*

*The initial state for a planning instance is $(0, Init_{logical}, \vec{x})$ where $\vec{x}$ is the vector of values in $\mathbb{R}_\perp$ corresponding to the initial assignments given by $Init_{numeric}$ (treating unspecified values as $\perp$).*

**Core Definition 3 Assignment Proposition** *The syntactic form of a numeric effect consists of an assignment operator (`assign`, `increase`, `decrease`, `scale-up` or `scale-down`), one primitive numeric expression, referred to as the lvalue, and a numeric expression (which is an arithmetic expression whose terms are numbers and primitive numeric expressions), referred to as the rvalue.*

*The* assignment proposition *corresponding to a numeric effect is formed by replacing the assignment operator with its equivalent arithmetic operation (that is* (`increase p q`) *becomes* (`= p (+ p q)`) *and so on) and then annotating the lvalue with a "prime".*

*A numeric effect in which the assignment operator is either* `increase` *or* `decrease` *is called an* additive assignment effect, *one in which the operator is either* `scale-up` *or* `scale-down` *is called a* scaling assignment effect *and all others are called* simple assignment effects.

**Core Definition 4 Normalisation** *Let I be a planning instance of dimension $dim_I$ and let*

$$index_I \ : \ PNEs_I \ \rightarrow \ \{1, \dots \ , dim\}$$

*be an (instance-dependent) correspondence between the primitive numeric expressions and integer indices into the elements of a vector of $dim_I$ real values, $\mathbb{R}^{dim_I}_{\perp}$.*

*The normalised form of a ground proposition, p, in I is defined to be the result of substituting for each primitive numeric expression f in p, the literal $X_{index_I(f)}$. The normalised form of p will be referred to as $\mathcal{N}(p)$. Numeric effects are normalised by first converting them into assignment propositions. Primed primitive numeric expressions are replaced with their corresponding primed literals. $\vec{X}$ is used to represent the vector $\langle X_1 \dots X_n \rangle$.*

**Core Definition 5 Flattening Actions** *Given a planning instance, I, containing an action schema $A \in As_I$, the set of action schemas flatten(A), is defined to be the set S, initially containing A and constructed as follows:*

- *While S contains an action schema, X, with a conditional effect,* (`when P Q`)*, create two new schemas which are copies of X, but without this conditional effect, and conjoin the condition* `P` *to the precondition of one copy and* `Q` *to the effects of that copy, and conjoin* (`not P`) *to the precondition of the other copy. Add the modified copies to S.*

- *While S contains an action schema, X, with a formula containing a quantifier, replace X with a version in which the quantified formula* ( `Q` ( $var_1 \dots var_k$ ) `P`) *in X is replaced with the conjunction (if the quantifier, Q, is* `forall`*) or disjunction (if Q is* `exists`*) of the propositions formed by substituting objects in I for each variable in $var_1 \dots var_k$ in* `P` *in all possible ways.*

*These steps are repeated until neither step is applicable.*

**Core Definition 6 Ground Action** *Given a planning instance, I, containing an action schema $A \in As_I$, the set of ground actions for A, $GA_A$, is defined to be the set of all the structures, a, formed by substituting objects for each of the schema variables in each schema, X, in flatten(A) where the components of a are:*

- Name *is the name from the action schema, X, together with the values substituted for the parameters of X in forming a.*

- $Pre_a$, *the* precondition *of a, is the propositional precondition of a. The set of ground atoms that appear in $Pre_a$ is referred to as $GPre_a$.*

- $Add_a$, *the* positive postcondition *of a, is the set of ground atoms that are asserted as positive literals in the effect of a.*

- $\text{Del}_a$, *the* negative postcondition *of $a$,is the set of ground atoms that are asserted as negative literals in the effect of $a$.*

- $\text{NP}_a$, *the* numeric postcondition *of $a$, is the set of all assignment propositions corresponding to the numeric effects of $a$.*

*The following sets of primitive numeric expressions are defined for each ground action, $a \in GA_A$:*

- $L_a = \{f \,|\, f$ *appears as an lvalue in $a\}$*

- $R_a = \{f \,|\, f$ *is a PNE in an rvalue in $a$ or appears in $Pre_a\}$*

- $L_a^* = \{f \,|\, f$ *appears as an lvalue in an additive assignment effect in $a\}$*

**Core Definition 7 Valid Ground Action** *Let $a$ be a ground action. $a$ is* valid *if no primitive numeric expression appears as an lvalue in more than one simple assignment effect, or in more than one different type of assignment effect.*

**Core Definition 8 Updating Function** *Let $a$ be a valid ground action. The* updating function *for $a$ is the composition of the set of functions:*

$$\{\text{NPF}_p \;:\; \mathbb{R}_\perp^{dim} \to \; \mathbb{R}_\perp^{dim} \,|\, p \in NP_a\}$$

*such that $\text{NPF}_p(\vec{x}) = \vec{x}'$ where for each primitive numeric expression $x_i'$ that does not appear as an lvalue in $\mathcal{N}(p)$, $x_i' = x_i$ and $\mathcal{N}(p)[\vec{X}' := \vec{x}', \vec{X} := \vec{x}]$ is satisfied.*

*The notation $\mathcal{N}(p)[\vec{X}' := \vec{x}', \vec{X} := \vec{x}]$ should be read as the result of normalising $p$ and then substituting the vector of actual values $\vec{x}'$ for the parameters $\vec{X}'$ and actual values $\vec{x}$ for formal parameters $\vec{X}$.*

**Core Definition 9 Satisfaction of Propositions** *Given a logical state, $s$, a ground propositional formula of* PDDL2.1, *$p$, defines a predicate on $\mathbb{R}_\perp^{dim}$, $Num(s, p)$, as follows:*

$$Num(s, p)(\vec{x}) \quad iff \quad s \models \mathcal{N}(p)[\vec{X} := \vec{x}]$$

*where $s \models q$ means that $q$ is true under the interpretation in which each atom, $a$, that is not a numeric comparison, is assigned true iff $a \in s$, each numeric comparison is interpreted using standard equality and ordering for reals and logical connectives are given their usual interpretations. $p$ is* satisfied *in a state $(t, s, \vec{x})$ if $\text{Num}(s, p)(\vec{x})$.*

*Comparisons involving $\perp$, including direct equality between two $\perp$ values are all undefined, so that enclosing propositions are also undefined and not satisfied in any state.*

**Core Definition 10 Applicability of an Action** *Let $a$ be a ground action. $a$ is* applicable *in a state $s$ if the $Pre_a$ is satisfied in $s$.*

## Appendix B. Proof of theorem

**Theorem 2** PDDL+$_{PTA}$ *has indirectly equivalent expressive power to Priced Timed Automata.*

We begin by showing that an arbitrary PTA can be expressed as a PDDL+$_{PTA}$ domain without any blow-up in the size of the encoding. We then show that the converse is also the case.

Given a PTA $\langle L, l_0, E, I, P \rangle$ we construct a PDDL+$_{PTA}$ model as follows. For each edge $e = (l_i, g, a, r, l_j)$, where $g$ is the clock constraint, $a$ is the action and $r$ is the subset of clock variables reset by this edge, we construct the following instantaneous action schema, called a *transition action*, which models the instantaneous transition from $l_i$ into $l_j$.

```
(:action transitione
 :parameters ()
 :precondition (and (in li) g)
 :effect (and {∀x ∈ r · (assign (x) 0)}
              (increase (c) P(e))
              (not (in li))
              (in lj)))
```

We also construct, for each location $l_i$, the following process schema:

```
(:process process-locationli
 :parameters ()
 :precondition (in li)
 :effect       (and (increase (c) (* #t P(li))
                    {∀x ∈ C · (increase (x) (* #t 1))}
                    ))
```

Finally, we construct an event for each location:

```
(:event event-locationli
 :parameters ()
 :precondition (and (in li) (not (I(li))))
 :effect       (not (in li)))
```

The initial state specifies that each clock variable and the cost variable starts with value 0. It asserts (`in l`$_0$). The special error state is the empty state (if an event is triggered then it will remove the current location proposition, leaving it impossible to progress).

Note that this domain is valid PDDL+$_{PTA}$. No two actions can be applied in parallel because the system can only ever satisfy one condition of the form (`in l`$_i$).

For this construction to correctly capture the PTA, it remains to be shown that any PTA trajectory corresponds to a PDDL+$_{PTA}$ plan for this domain and that any plan corresponds to a trajectory.

Consider a (valid) PTA trajectory,

$$(l_0, u_0) \xrightarrow{\delta_1} (l_1, u_1) \xrightarrow{\delta_2} \dots (l_n, u_n)$$

where each $u_i$ is a clock valuation and each $\delta_i$ is a transition which may either be an edge or a positive time delay. In the case of time delay transitions, the location remains the same but the clock valuation is updated by the delay duration, while for edge transitions the location is updated but the clock valuation remains the same. This trajectory is mapped to a plan by creating an action instance for each edge transition (using the corresponding action in the PDDL+$_{PTA}$ description). The action is set to be applied at the time corresponding to the sum of the time delay transitions that precede the transition in the trajectory. This is a valid plan since the processes are active at exactly the same times in the PDDL model as the corresponding time transitions in the trajectory.

Similarly, given any valid PDDL plan for the above domain, the plan defines a trajectory by mapping the instantaneous actions into their corresponding edge transitions and the gaps between them into time delay transitions. The trajectory will be a valid one because of the construction of the actions and process models. Note that no valid plan can trigger an event, since this will leave the system in a state that cannot be progressed, preventing satisfaction of any goal. This means that invariant conditions for each location are always maintained.

We now consider the opposite direction of the proof: PDDL+$_{PTA}$ planning instances yield state space and transition models that can be expressed as PTAs with a constant factor transformation. We observe that the constraints on the PDDL+$_{PTA}$ language ensures that clock variables and a cost variable are distinguished and behave exactly as required for a PTA. In the ground state space, all states correspond directly to locations of a PTA and the legal actions that transition between states correspond to the edges of the corresponding PTA. With the exception of event transitions leading into an exceptional error state, each edge in a PDDL+$_{PTA}$ transition system will, by the constraints defining the language, always correspond to an instantaneous action and this action determines the clock constraints and reset effects of the corresponding PTA edge. Exactly one process is active in each state, and causes the variables to behave as clocks (except for the cost variable). The events govern the invariants for the states, causing any violation of an invariant condition to trigger a transition into an error state.

The correspondence between trajectories for the PTA defined in this way and plans for the PDDL+$_{PTA}$ transition system is immediate, subject to the same observations made above.

$\square$

## Appendix C. The Planetary Lander Domain in PDDL+

The following domain encoding shows the PDDL+ model for the Planetary Lander problem discussed in Section 2.2. According to implementation details of systems designed to handle PDDL+ models, it might be necessary to introduce further events to manage the transition between charging and discharging, because imprecision of the system in measuring the values of the demand and supply might lead to difficulties at the boundary where supply and demand are equal.

```
(define (domain power)
 (:requirements :typing :durative-actions :fluents :time
```

```
                       :negative-preconditions :timed-initial-literals)
 (:types equipment)
 (:constants unit - equipment)
 (:predicates (day) (commsOpen) (readyForObs1) (readyForObs2)
                     (gotObs1) (gotObs2)
                     (available ?e - equipment))
 (:functions (demand) (supply) (soc) (charge-rate) (daytime)
             (heater-rate) (dusk) (dawn)
             (fullTime) (partTime1) (partTime2)
             (obs1Time) (obs2Time) (obs1-rate) (obs2-rate)
             (A-rate) (B-rate) (C-rate) (D-rate) (safeLevel)
             (solar-const))

(:process charging
 :parameters ()
 :precondition (and (< (demand) (supply)) (day))
 :effect (and (increase (soc) (* #t (* (* (- (supply) (demand))
                                          (charge-rate))
                                     (- 100 (soc)))
                              )))
)

(:process discharging
 :parameters ()
 :precondition (> (demand) (supply))
 :effect (decrease soc (* #t (- (demand) (supply)))))
)

(:process generating
 :parameters ()
 :precondition (day)
 :effect (and (increase (supply)
                     (* #t (* (* (solar-const) (daytime))
                              (+ (* (daytime)
                                    (- (* 4 (daytime)) 90)) 450))))
              (increase (daytime) (* #t 1)))
)

(:process night-operations
 :parameters ()
 :precondition (not (day))
 :effect (and (increase (daytime) (* #t 1))
              (decrease (soc) (* #t (heater-rate))))
)

(:event nightfall
 :parameters ()
 :precondition (and (day) (>= (daytime) (dusk)))
 :effect (and (assign (daytime) (- (dawn)))
              (not (day)))
```

```
)

(:event daybreak
 :parameters ()
 :precondition (and (not (day)) (>= (daytime) 0))
 :effect (day)
)

(:durative-action fullPrepare
 :parameters ()
 :duration (= ?duration (fullTime))
 :condition (and (at start (available unit))
                 (over all (> (soc) (safelevel))))
 :effect (and (at start (not (available unit)))
              (at start (increase (demand) (A-rate)))
              (at end (available unit))
              (at end (decrease (demand) (A-rate)))
              (at end (readyForObs1))
              (at end (readyForObs2)))
)

(:durative-action prepareObs1
 :parameters ()
 :duration (= ?duration (partTime1))
 :condition (and (at start (available unit))
                 (over all (> (soc) (safelevel))))
 :effect (and (at start (not (available unit)))
              (at start (increase (demand) (B-rate)))
              (at end (available unit))
              (at end (decrease (demand) (B-rate)))
              (at end (readyForObs1)))
)

(:durative-action prepareObs2
 :parameters ()
 :duration (= ?duration (partTime2))
 :condition (and (at start (available unit))
                 (over all (> (soc) (safelevel))))
 :effect (and (at start (not (available unit)))
              (at start (increase (demand) (C-rate)))
              (at end (available unit))
              (at end (decrease (demand) (C-rate)))
              (at end (readyForObs2)))
)

(:durative-action observe1
 :parameters ()
 :duration (= ?duration (obs1Time))
 :condition (and (at start (available unit))
                 (at start (readyForObs1))
                 (over all (> (soc) (safelevel))))
```

```
                      (over all (not (commsOpen))))
 :effect (and (at start (not (available unit)))
              (at start (increase (demand) (obs1-rate)))
              (at end (available unit))
              (at end (decrease (demand) (obs1-rate)))
              (at end (not (readyForObs1)))
              (at end (gotObs1)))
)

(:durative-action observe2
 :parameters ()
 :duration (= ?duration (obs2Time))
 :condition (and (at start (available unit))
                 (at start (readyForObs2))
                 (over all (> (soc) (safelevel)))
                 (over all (not (commsOpen))))
 :effect (and (at start (not (available unit)))
              (at start (increase (demand) (obs2-rate)))
              (at end (available unit))
              (at end (decrease (demand) (obs2-rate)))
              (at end (not (readyForObs2)))
              (at end (gotObs2)))
)
)
```

## Appendix D. Durative Actions in PDDL+

The syntax for durative actions can be seen as a purely syntactic convenience. In order to support this view, durative actions must be mapped directly into an equivalent start-process-stop representation using the basic syntax of PDDL+. The reason for performing the mapping is in order to give durative actions a semantics in terms of the underlying structures of PDDL+, which are themselves given meaning in terms of hybrid automata as discussed in Section 6. The mapping is as follows.

Consider the following generic structure for a PDDL2.1 durative action, excluding the use of duration inequalities and continuous effects. Conditional effects are ignored, since they can be handled by flattening the actions as described in Core Definition 5. The complication of conditional effects that combine initial and final conditions is discussed in (Fox & Long, 2003) and does not affect the basic principles we demonstrate in the translation we give here. Similarly, for convenience, we do not consider duration constraints referring to the end state — extension of the treatment below to manage these constraints is straightforward.

```
(:durative-action name
     :parameters (p⃗)
     :duration (= ?duration Dur[p⃗])
     :condition (and (at start Pre_S) (at end Pre_E) (over all Inv))
     :effect (and (at start Post_S)(at end Post_E[?duration]))
)
```

We now construct the following structures in PDDL+:

```
(:action name-start
     :parameters (p⃗)
     :precondition (and Pre_S (not (name_clock_started p⃗)))
     :effect (and Post_S
                     (name_clock_started p⃗)
                     (assign (name_clock p⃗) 0)
                     (assign (name_duration p⃗) Dur[p⃗])
                     (increase (clock_count) 1))

(:process name-process
     :parameters (p⃗)
     :precondition (name_clock_started p⃗)
     :effect (increase (name_clock p⃗) (* #t 1))

(:event name-failure
     :parameters (p⃗)
     :precondition (and (name_clock_started p⃗)
                     (not (= (name_clock p⃗) (name_duration p⃗)))
                     (not Inv))
     :effect (assign (name_clock p⃗) (+ (name_duration p⃗) 1)))

(:action name-end
     :parameters (p⃗)
     :precondition (and Pre_E (name_clock_started p⃗)
                     (= (name_clock p⃗) (name_duration p⃗)))
     :effect (and Post_E[(name_duration p⃗)]
                     (not (name_clock_started p⃗))
                     (decrease (clock_count) 1)))
```

To complete the transformation, the initial state has the (= (clock_count) 0) added to it and the goal has the condition (= (clock_count) 0) added to it.

Note that the clock is uniquely determined by the name and arguments of the durative action, so the clock is not shared between different durative actions. Also note that there is only one action that can start each such clock and only one that can terminate it. If a plan makes use of the durative action then in the transformed domain the durative action can be simulated by the actions that start and stop the clock. In order for these actions to execute successfully the conditions must be identical to those stipulated in the durative action and their effects are equivalent to the original durative action. Furthermore, the end action can only be executed once the clock has reached the correct duration value (which is recorded at the start). The clock is driven by a process that is active only once the start action has been executed and only until the end action is executed. An event is used to monitor the invariant conditions. If ever the invariant becomes false while the clock is running then the event sets the clock to be after the duration of the durative action. This makes it impossible to complete the action and terminate the clock, so that no valid execution trace can be constructed to achieve the goal in which the event is triggered. Every time a clock starts a count is incremented and it is decremented when a clock stops. The count must be zero at the end, meaning that every clock must have stopped and therefore every durative action must have been ended before the plan is complete.

The duration of the action is managed by using a metric fluent to store the duration at the outset in order to use the value at the conclusion of the action. If the value is a fixed constant then this can be simplified by replacing the metric fluent with the appropriate constant.

Packaging actions, events and processes into a durative action abstracts the details of how the interval ends — whether by an action or by an event. For example, if a durative action is used to represent the activity of filling a bath, its end point will represent the action of turning off the taps, whilst in a durative action representing a ball being dropped, the end of the action is the event of the ball landing. However, there is no syntactic difference between the structures encoding these examples that allows a distinction to be drawn between an end point that is an action and one that is an event. The mapping of durative actions into actions, processes and events requires an arbitrary decision to be made about how to handle the end points of durative actions. While we have chosen to use actions, an alternative formulation is possible using events.

The mapping converts every durative action into the family of PDDL+ constructs shown above — two actions to start and end the interval, a process to execute during the interval, and a monitoring event for the invariant. Because an action is always used to end the interval a planner must always choose to apply it in order to obtain a goal state. In the final plan the terminating action will appear, chosen by the planner, even though its application is in fact forced, and its point of application must be consistent with the duration constraint of the original durative action. Thus, although the planner is free to choose whether and when to apply the action, in order to construct a valid plan it is forced to apply the action at a point that exactly meets the appropriate temporal constraints. Of course, the end points of the simulated durative actions can be trivially post-processed to make a plan containing durative actions instead of to their underlying components.